\pdfoutput=1

\documentclass[11pt]{article}

\normalsize 
\usepackage[final]{acl}

\usepackage{times}
\usepackage{latexsym}
\usepackage{amstext}

\usepackage[T1]{fontenc}

\usepackage[utf8]{inputenc}

\usepackage{microtype}

\usepackage{inconsolata}

\usepackage{adjustbox}
\usepackage{float}
\usepackage{algcompatible}
\usepackage[linesnumbered,ruled,vlined]{algorithm2e}
\usepackage{acl}
\usepackage{graphicx}
\usepackage{multirow}
\usepackage{amsmath}
\usepackage{booktabs}
\usepackage{tabularx}
\usepackage{bm}
\usepackage{tgheros}
\usepackage{amssymb}
\usepackage{svg} 
\usepackage{booktabs}
\usepackage{multirow}
\usepackage{longtable}
\usepackage{caption}
\usepackage{placeins}
\usepackage{afterpage}
\usepackage{needspace}
\usepackage{lipsum} 
\usepackage{float}  
\usepackage{afterpage} 
\newcommand{\citeauthorandyear}[1]{\citeauthor{#1} (\citeyear{#1})}
\usepackage{ulem}
\newcommand{\softmax}{\operatorname{softmax}}
\title{Retrieval Visual Contrastive Decoding to Mitigate Object Hallucinations in Large Vision-Language Models}

\author{
  Jihoon Lee \\
  Yonsei University \\
  Onoma AI \\
  \texttt{jihoonlee98@yonsei.ac.kr}
  \And
  Min Song\thanks{Corresponding author.} \\
  Yonsei University \\
  Onoma AI \\
  \texttt{min.song@onomaai.com}
}

\begin{document}
\twocolumn 
\sloppy
\vfill
\maketitle
\begin{abstract}
Despite significant advancements in Large Vision-Language Models, Object Hallucination (OH) remains a persistent challenge. Building upon prior studies on contrastive decoding that address this issue without requiring additional model training, we introduce RVCD (Retrieval Visual Contrastive Decoding), an advanced method to suppress OH. RVCD leverages both negative and positive images at the logit level, explicitly referencing AI-generated images designed to represent a single concept. Our approach demonstrates substantial improvements over existing decoding-based methods. Code and data are released at \href{https://github.com/JiHoonLee9898/RVCD}{\url{https://github.com/JiHoonLee9898/RVCD}}.
\end{abstract}

\section{Introduction}

Large Vision Language Models (LVLMs) are models designed to generate sophisticated textual responses based on multimodal inputs of images and text. In recent years, successful experiments on integrating vision encoders with language models have demonstrated promising progress in this field~\cite{zhu2023minigpt4, liu2023visual, zhang2024rankclip}.

However, Large Vision Language Models (LVLMs) are still not free from the issue of Object Hallucination (OH), which refers to the phenomenon where LVLMs erroneously generate hallucinated objects and descriptions in their outputs~\cite{rohrbach2018object}.

OH can be categorized into three types: generating descriptions of objects that do not exist in the image (existence), misdescribing the attributes of existing objects (attribute), and incorrectly describing the relationships between objects (relationship)~\cite{gunjal2023detecting, zhai2023halle}.

Previous studies have demonstrated that even more sophisticated and larger LVLMs are not free from the issue of Object Hallucination (OH)~\cite{dai2022plausible, li2023evaluating, guan2023hallusionbench}.

To address Object Hallucination (OH), which undermines the reliability of LVLMs, various methodologies have been proposed. These include approaches that mitigate OH by modifying the outputs generated by LVLMs~\cite{zhou2023analyzing}, introducing self-correction pipelines~\cite{yin2023woodpecker}, or employing decoding-based methods~\cite{huang2023opera, leng2023visual, chen2024halc, zhuang2024game}.

Among these methodologies, visual contrastive decoding-based approaches are particularly attractive and practical because they mitigate OH without requiring additional training for the models~\cite{leng2023visual, chen2024halc, zhuang2024game}. These methods distort the input source images~\cite{leng2023visual}, or zoom into the local views containing important objects in the source images~\cite{chen2024halc, zhuang2024game} to generate logits for regulation. Theses logits modify or replace the logits generated from the original input images, contributing to producing better outputs.

However, despite the excellence of their methods, they do not fully exploit the potential of visual contrastive decoding — the potential that the images used to generate logits for regulation do not always need to be transformations of the original images. 

We introduce a novel method called RVCD (Retrieval-Visual Contrastive Decoding), which maximizes the regulation strength by retrieving and leveraging multiple explicit images as regulatory targets. The explicit images we use are designed to encapsulate a \textit{single concept}, enabling the contrastive decoding process to add or subtract the desired or undesired concept effectively, thereby allowing the adjusted logits to clearly align with the target image.

Our method retrieves multiple explicit reference images, generates negative logits to be regulated, and recovers positive logits lost during the regulation process, utilizing them at every decoding step for token generation. These explicit reference images are created using image generation models to represent single concepts and are ultimately selected based on agreement conditions between the image generation models and LVLMs.
Similar to prior studies, our method can be easily applied to open-source LVLMs, such as MiniGPT-4~\cite{chen2023minigptv2}, LLaVA~\cite{liu2023visual}, and mPLUG-Owl2~\cite{ye2023mplugowl2}. 

Our contributions are summarized as follows: 
(1) We propose a novel plug-and-play decoding method called RVCD (Retrieval-Visual Contrastive Decoding). This method is train-free, strongly regulates OH, and simultaneously preserves high output text quality.
(2) We provide a retrieval database utilized in RVCD. All images in the database are aligned with the consensus between diffusion-based image generation models and LVLMs, and they represent single concepts. This database enhances the explainability of our RVCD and can serve as a resource for future research leveraging explicit concepts in visual contrastive decoding.
(3) Through comprehensive experiments, we demonstrate the strong OH reduction capability of RVCD, which significantly outperforms existing methods. 

\section{Related Work}
Object Hallucination (OH) refers to the phenomenon where BERT-based Vision Language Models (VLMs)~\cite{li2019visualbert, radford2021learning}, or more recent LVLMs~\cite{liu2023visual, zhu2023minigpt4, tu2023unicorns, cui2023holistic, wang2024mementos, zhou2024calibrated}, erroneously generate unfaithful contents.
\citeauthorandyear{gunjal2023detecting} and \citeauthorandyear{zhai2023halle} categorized OH into three types: existence, attribute, and relationship OH. These correspond to generating descriptions of non-existent objects, producing misleading descriptions, and generating incorrect descriptions of relationships between existing objects, respectively.

The most dominant metric for evaluating OH is CHAIR~\cite{rohrbach2018object}. This metric can be used in scenarios where a finite synonym dictionary and a set of ground truth objects mapped to an image are defined. It evaluates the proportion of synonyms appearing in LVLM outputs that are not defined in the ground truth object set (CHAIR$_I$), and the proportion of sentences where CHAIR$_I$ is non-zero (CHAIR$_S$).
Another well-known recent metric is POPE~\cite{li2023evaluating}, which measures the degree of OH using precision, recall, and accuracy. To do so, POPE frames a binary classification problem for the LVLM, evaluating its outputs based on the inclusion of positive or negative assertions (e.g., "yes" or "no").
Additionally, the traditional and standard text generation quality metric BLEU~\cite{papineni2002bleu} is still utilized in recent studies~\cite{chen2024halc, zhuang2024game}. BLEU serves as an additional indicator to ensure that little sacrifice in text quality occurs while mitigating OH~\cite{chen2024halc}.

Efforts to mitigate Object Hallucination (OH) have been ongoing since the introduction of the CHAIR metric by \citeauthorandyear{rohrbach2018object}, yet OH remains an unsolved challenge despite advancements in LVLMs~\cite{dai2022plausible, li2023evaluating, zhou2024aligning}. No LVLM to date has completely resolved OH in its outputs. To address OH, recent approaches have explored various strategies.
\citeauthorandyear{sun2023RLHF} and \citeauthorandyear{jing2024fgaif} proposed reinforcement learning-based methods for training model parameters, while \citeauthorandyear{xing2024cca} introduced a token reordering approach to mitigate the issue of long-term decay in Rotary Position Encoding (RoPE).
\citeauthorandyear{zhou2023analyzing} and~\citeauthorandyear{yin2023woodpecker} proposed post-hoc or self-correction pipelines to reduce OH in final text outputs. 
Meanwhile,~\citeauthorandyear{huang2023opera},~\citeauthorandyear{leng2023visual},~\citeauthorandyear{chen2024halc}, ~\citeauthorandyear{zhuang2024game}, and~\citeauthorandyear{yang2024pensieve} introduced decoding based strategies. These approaches demonstrated that adjusting logits during the decoding steps, without training or modifying output text directly, can effectively reduce OH.

However, despite the excellence of the core idea of visual contrastive decoding (VCD)~\cite{leng2023visual}, including the above studies that leverage it, they fail to fully exploit its hidden potential. Specifically, they overlook the fact that the types of images used to generate regulated logits are not necessarily restricted to variations of the original input image.

Building on the intuition that explicit images for regulation can be derived from the external database, we propose Retrieval-Visual Contrastive Decoding (RVCD). Specifically, we constructed a database by generating explicit AI-created images that best represent the single concept of each word in the finite vocabulary used for OH evaluation~\cite{rohrbach2018object}. In our approach, multiple regulated and preserved logits are generated from images retrieved from this database.

\section{Background and Motivation}

\subsection{Problem Definition}

A typical LVLM, parameterized by $\theta$, encodes an input text query $x$ and an input image $v$, integrates the encoded embeddings to generate a multimodal embedding, and processes it autoregressively:
\begin{equation}
y_t \sim p_\theta(\cdot \mid v, x, y_{<t}) \propto \exp f_\theta(\cdot \mid v, x, y_{<t}),
\end{equation}
where $y_t$ represents the token of the time step ($t$), $y_{<t}$ is the sequence of output tokens generated up to the time step ($t-1$), and $f_\theta$ is the logit distribution (unnormalized log-probabilities) generated by the LVLM ($\mathcal{M}_{\theta}^{\mathrm{LVLM}}$).

Object Hallucination (OH) occurs when the information from the input image $v$ conflicts with some tokens in $y$. To mitigate OH, $y$ must describe only the information present in $v$ while maintaining high text generation quality.

\subsection{Our Approach to OH Mitigation}
\begin{figure}[t]
    \centering
    \includegraphics[width=0.485\textwidth]{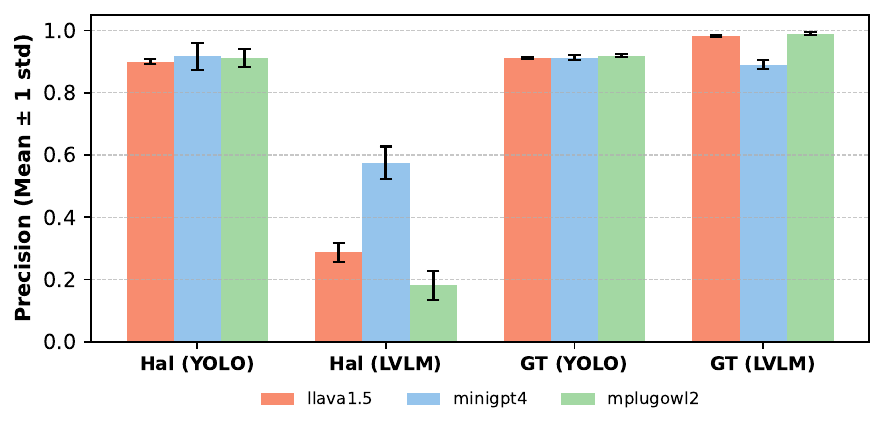} 
    \captionsetup{aboveskip=5pt, belowskip=-10pt}
    \caption{Detection precision for YOLO and LVLM detectors on MSCOCO Validation 2014~\cite{lin2014mscoco}. Hal (\(\cdot\)) shows the proportion of hallucinated objects from greedy-decoded captions detected by YOLO and LVLMs VQA that were true hallucinations. GT (\(\cdot\)) illustrates the proportion of objects correctly identified as existing by YOLO and LVLMs. While both perform similarly in detecting existing objects, YOLO excels in hallucination detection, motivating us to transfer YOLO’s strength to LVLMs for correcting hallucinated objects. The statistical details are provided in Appendix~\ref{appendix_precision}.}
    \label{tab:detector_bar}
\end{figure}

We propose an approach to mitigate Object Hallucination (OH) by detecting hallucinated objects that should not have been generated in the greedy-decoded draft output and regulating this information to produce the target output. To this end, we first conducted an experiment to assess whether LVLMs can self-check OH occurrences in their draft outputs. However, as shown in Figure~\ref{tab:detector_bar}, object detection capability of LVLMs is insufficient to accurately detect hallucinated objects in their draft captions.

On the other hand, traditional object detection (OD) model YOLO~\cite{redmon2015yolo}, which lack linguistic capabilities, shows much better hallucination detection precision than LVLMs VQA (Vision Question Answering) outputs. In this work, we utilized YOLOv8x~\cite{yolov8} due to its significant influence and widespread adoption in the deep learning community for both training and inference~\cite{yolov8_blog}.

We hypothesize that providing accurate OH detection information from the OD model at each decoding step of the LVLM will suppress hallucinated tokens while maintaining fluent language generation. To achieve this, we generate multiple logits from retrieved explicit images based on the OH detection information from the OD model and regulate them at each decoding step to mitigate OH.

\textbf{Our Goal:} Conveying accurate OH detection information of OD models to the LVLM's token generation stage to minimize OH in the target output while maintaining fluent language generation capability.


\begin{figure*}[ht]
    \captionsetup{aboveskip=-15pt, belowskip=-10pt} 
    \centering
    \includegraphics[width=\textwidth]{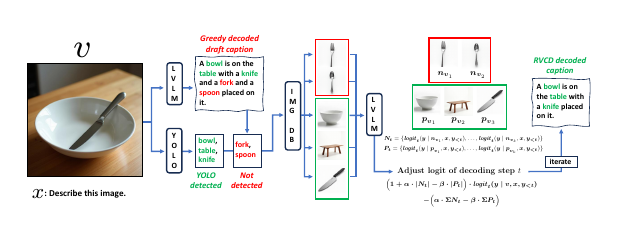}
    \caption{Overall pipeline of our RVCD. $x$ denotes the input prompt, and $v$ denotes the input image. $n_{vi}$ and $p_{vi}$ are images retrieved from the image database, representing single-concept images for objects identified as hallucinations (appearing only in the draft caption) and ground truth (appearing in both the OD model and draft caption), respectively. $N_t$ and $P_t$ represent the sets of logits generated from $n_{vi}$ and $p_{vi}$, respectively. At each decoding step, the LVLM processes $x$, $v$, $N_t$, $P_t$, and ongoing output tokens $y_{<t}$, which are then integrated according to our proposed formula. This iterative decoding process produces the final caption of RVCD.}
    \label{fig:main_figure}
\end{figure*}

\section{Methodology}
The overview of our method is illustrated in Figure~\ref{fig:main_figure}.

First, we generate a greedy-decoded text result for the input image using the LVLM, which we refer to as the draft caption. Simultaneously, we obtain a list of objects present in the same image using an object detection (OD) model such as YOLO. 

Objects mentioned in the draft caption but not detected by the OD model are classified as negative objects.
Objects detected by both the draft caption and the OD model are classified as positive objects.

Our goal is to suppress the generation of tokens related to negative objects while preserving the representation of positive objects at every token generation step, by corresponding logits generated from retrieved explicit images. We describe the details of each step below.

\begin{figure}[t] 
    \centering
    \includegraphics[width=0.41\textwidth]{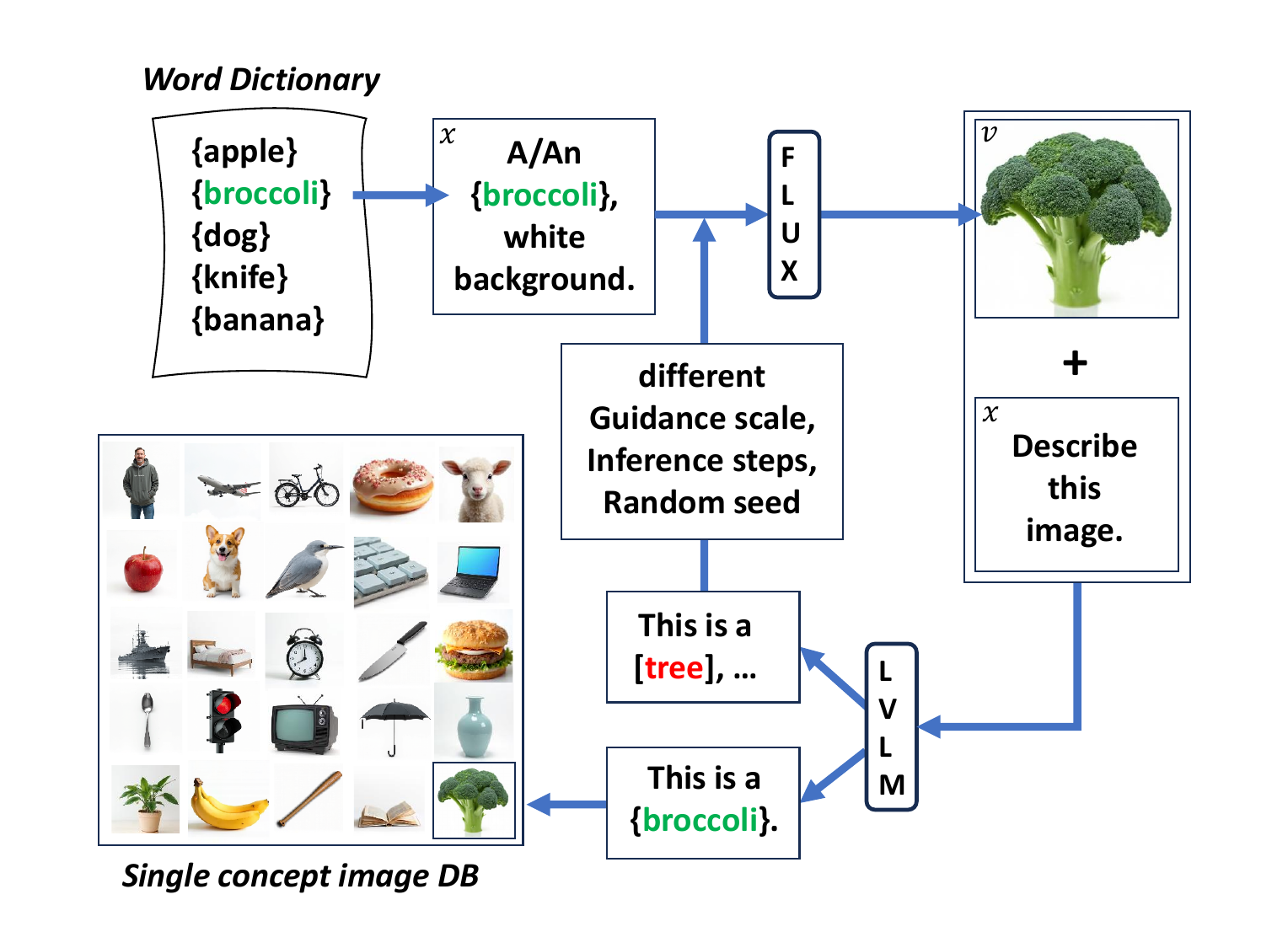}
    \captionsetup{aboveskip=5pt, belowskip=-15pt}
    \caption{AI generated single concept image DB. we adopted FLUX.1-dev~\cite{yang2024flux} to generate $336$ * $336$ pixels images representing only a single concept corresponding to each word in the MSCOCO objects synonyms dictionary and stored them in an image database. Images were stored in the database only if the LVLM’s output captions and the image generation model’s input prompts both mentioned the corresponding concept. Otherwise, the hyperparameters of the image generation model were adjusted, and the images were regenerated. This process was repeated until images were generated for every word in the dictionary.}

    \label{fig:Ai generated single concept image DB}
\end{figure}

\subsection{Generate Reference Images Corresponding to CHAIR Dictionary Words}
For the quantitative evaluation of OH in output captions, prior studies have utilized the dictionary of CHAIR~\cite{rohrbach2018object}. This dictionary is used to extract a list of objects from output descriptions via natural language processing (NLP) and compare it against a ground truth list. We extend this dictionary into the visual domain. Specifically, for every words (over 400) in the dictionary including the representative terms of 80 MSCOCO objects and their synonyms, we generated AI images that represent only the corresponding concept and mapped them to their respective entries as Figure~\ref{fig:Ai generated single concept image DB}.

We adopted FLUX.1-dev~\cite{yang2024flux} as the image generation model and used the prompt “An/A \{object\}, white background” to create images that exclusively represent the single concept of the object. These images were then re-fed to the LVLM along with the prompt “Describe this image in detail.” We adopted llava-1.5~\cite{liu2023visual} as the LVLM. The image is stored only if its caption included the \{object\}, indicating alignment between the intended prompt of the image generation model and the interpretation by the LVLM.

The extended CHAIR dictionary, which includes an image database mapped to every word, serves as a reference database that retrieves a corresponding reference image visually representing the single concept associated with the word.

\subsection{Comparison Between Draft Caption and OD Detected Object List}
The base equation of general greedy decoding is as follows:
\begin{equation}
y_t = \arg\max( \softmax[f_\theta(\cdot \mid v, x, y_{<t})]).
\end{equation}
Here, $f_\theta$ is the logit distribution generated by the LVLM ($\mathcal{M}_{\theta}^{\mathrm{LVLM}}$). Using greedy decoding with the LVLM, we obtain a draft caption for the input image and extract all mentioned objects to create a draft objects list.

Similarly, we use the OD model to generate a detected objects list for the same input image.
Duplicate objects in the draft objects list and the detected objects list are represented as a single unique object. Objects that exist in the draft objects list but not in the detected objects list are defined as $N$ (negative objects).
Objects that exist in both the draft objects list and the detected objects list are defined as $P$ (positive objects). This preprocessing step aims to exclude negative objects tokens while preserving positive objects tokens in the target output.

\subsection{Retrieval Visual Contrastive Decode}
For each object in $N$ and $P$, we retrieve a single corresponding image from the single concept image DB, and generate the output using the following formulas.
To incorporate the concepts of \( P_t \) and \( N_t \) at each decoding step \( t \):
\( P_t \) is a set of logits computed using images \( v_{p_i} \) from \( P \) (positive objects list), where each image is retrieved from the single concept image DB.
\[
P_t = \big\{ f_\theta(\cdot \mid v_{p_1}, x, y_{<t}), f_\theta(\cdot \mid v_{p_2}, x, y_{<t}), 
\]
\begin{equation}
            \dots, f_\theta(\cdot \mid v_{p_k}, x, y_{<t}) \big\}.
\end{equation}
\( N_t \) is a set of logits computed using images \( v_{n_i} \) from \( N \) (negative objects list), where each image is retrieved from the single concept image DB.
\[
N_t = \big\{ f_\theta(\cdot \mid v_{n_1}, x, y_{<t}), 
            f_\theta(\cdot \mid v_{n_2}, x, y_{<t}), 
\]
\begin{equation}
            \dots, f_\theta(\cdot \mid v_{n_m}, x, y_{<t}) \big\}.
\end{equation}
The adjusted logit at time \( t \) is defined as:
\[
f_{\text{adjusted}_t}(\cdot \mid v, x, y_{<t}) =
\]
\[
f_\theta(\cdot \mid v, x, y_{<t}) \cdot \Big( 1 + \alpha \cdot \text{len}(N) - \beta \cdot \text{len}(P) \Big)
\]
\begin{equation}
- \Big( \alpha \cdot \text{sum}(N_t) - \beta \cdot \text{sum}(P_t) \Big),
\label{tab:main_equation}
\end{equation}
by simplifying the following expression:
\[
f_{\text{adjusted}_t}(\cdot \mid v, x, y_{<t}) =
\]
\[
\mathcal{O}_{\text{logit}}
+ \alpha \left( \mathcal{O}_{\text{logit}} - N_{t_{1}} \right)
+ \cdots
+ \alpha \left( \mathcal{O}_{\text{logit}} - N_{t_{m}} \right)
\]
\begin{equation}
+ \beta \left( P_{t_{1}} - \mathcal{O}_{\text{logit}} \right)
+ \cdots
+ \beta \left( P_{t_{k}} - \mathcal{O}_{\text{logit}} \right),
\label{tab:main_equation}
\end{equation}
where $\mathcal{O}_{\text{logit}}$ denotes $f_\theta(\cdot \mid v, x, y_{<t})$ and $N_{t_{i}}$, $P_{t_{j}}$ are individual logits from the sets \( N_t \) and \( P_t \).
The logit distribution $f_{\text{adjusted}_t}(\cdot \mid \mkern-2.5mu v, x, y_{<t})$ is computed by the same model parameters \( \theta \), Using the negative and  positive logits, scaled by their respective parameters \( \alpha \) and \( \beta \). \( \text{len}(\cdot) \) represents the length of the list containing negative or positive images. Note that \( N \) and \( N_t \) have the same length, and \( P \) and \( P_t \) also have the same length. 
The final output token at decoding step $t$ is defined as:
\[
RVCD_{y_t} =
\]
\begin{equation}
\arg\max(\softmax[
f_{\text{adjusted}_t}(\cdot \mid v, x, y_{<t}]
).
\end{equation}
Here, the final output token index at decoding step $t$ is $RVCD_{y_t}$, which is obtained as the arg max from the softmax of \( f_{\text{adjusted}_t} \).

\subsection{Addressing Challenges with Negative Logits: Why \texorpdfstring{$\beta$}{beta} and Positive Logits?}
\label{tab:co_occur_section}
\begin{figure}[ht]
    \centering
    \begin{adjustbox}{center}
    \includegraphics[width=\linewidth]{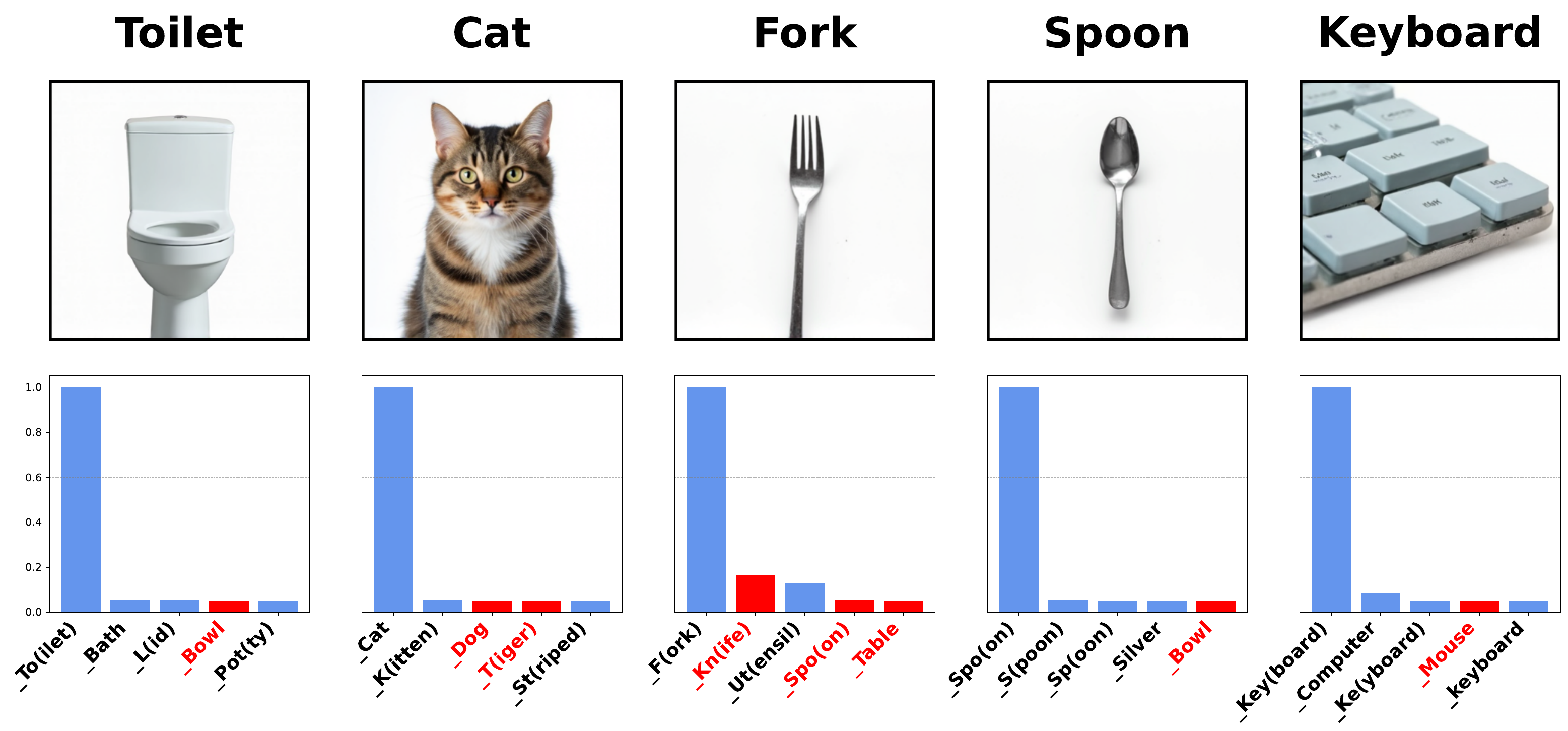}
    \end{adjustbox}
    \captionsetup{aboveskip=5pt, belowskip=-5pt}
    \caption{Top-5 token probabilities for each single concept images. When an LVLM is tasked with responding to an image using a single word, it frequently includes tokens representing other objects registered in the MSCOCO dictionary among its top-5 tokens, even for single-concept images. Details in Appendix~\ref{sec:appendixF}.} 
    \label{fig:top5_co_occur} 
\end{figure}
In RVCD, negative images represent a single concept. However, LVLMs often assign probabilities to tokens for commonly co-occurring objects~\cite{li2023evaluating, favero2024multi, chen2024multi} even when they are not explicitly present in the single concept image. For instance, with an image of a single fork, llava-1.5 ranks "fork" highest but still place "knife," "spoon," or "table" among the top predictions (Figure~\ref{fig:top5_co_occur}), due to their typical association in a kitchen setting.
This behavior introduces a risk in RVCD: subtracting multiple logits generated from negative images could unintentionally suppress representations of objects that are actually part of the ground truth. Removing these unintended yet valid objects can degrade caption quality.

To address this, we preserve ground truth object representations by reintroducing information from positive logits using a parameter \( \beta \). 
Positive logits are derived from reference images of objects identified by both the LVLM and the OD model. 
Our ablation study highlights the importance of 
\( \beta \).

\section{Experiments}
\textbf{Benchmarks.}
Following prior studies evaluating the performance of decoding methods~\cite{chen2024halc, zhuang2024game}, we assessed our RVCD using CHAIR~\cite{rohrbach2018object}, BLEU~\cite{papineni2002bleu}, and POPE~\cite{li2023evaluating} metrics on the MSCOCO dataset~\cite{lin2014mscoco}. Additionally, we conducted quantitative evaluations on the MME benchmark~\cite{fu2023mme}, and qualitative evaluation benchmark LLaVA-Bench~\cite{liu2023visual2}.
\newcommand{\OUTPUT}{\item[\textbf{Output:}]}
\begin{algorithm}[htbp]
\caption{RVCD Decoding}
\label{alg:adjusted_lvml_decoding}
\begin{algorithmic}[1]  
\REQUIRE  LVLM $\mathcal{M}_{\theta}^{\mathrm{LVLM}}$, text query $x$, image input $v$, object detection model $O_D$, image database \textit{imgDB}, finite word dictionary $D$.
\OUTPUT Adjusted RVCD decoded tokens $RVCDy_0$, \dots, $RVCDy_t$.
\STATE \textbf{Draft Decoding:}
\REPEAT
    \STATE For each decoding step, greedy decode:\\ 
    \hspace{0.75em}
    $y_t = \arg\max(\softmax[f_\theta(\cdot \mid v, x, y_{<t})])$
\UNTIL{Obtain decoded sequence $y_0, \dots, y_t$.}
\STATE Combine $y_0, \dots, y_t$ into \textit{draft}.

\STATE \textbf{Object Lists Generation:}
\STATE Extract words from \textit{draft} that are elements of $D$ to form a \textit{draft object list}.
\STATE Apply $O_D$ to input image $v$ to detect all objects and obtain \textit{OD object list}.

\STATE $\textit{draft object list} \gets \text{List}(\text{Set}(\textit{draft object list}))$
\STATE $\textit{OD object list} \gets \text{List}(\text{Set}(\textit{OD object list}))$

\STATE \textbf{Positive and Negative Object Pairing:}
\STATE $P \leftarrow []$, $N \leftarrow []$

\FOR{$o_i$ in \textit{draft object list}}
    \IF{$o_i$ $\in$ \textit{OD object list}}
        \STATE Append $($ \text{retrieved $v_{p_i}$ at \textit{imgDB}} \\
            \text{from $o_i$}$)$ to $P$.
    \ELSE
        \STATE Append $($ \text{retrieved $v_{n_i}$ at \textit{imgDB}} \\
            \text{from $o_i$}$)$ to $N$.
    \ENDIF
\ENDFOR

\STATE \textbf{RVCD Adjusted Decoding:}
\REPEAT
    \STATE For each decoding step $t$:
    \STATE RVCD decode with input $v$, $x$, $N_t$, $P_t$, $y_{<t}$.
    \STATE \textbf{Definitions:}
    \STATE \textit{$N_t$}: $\left\{f_\theta(\cdot \mid v_{n_i}, x, y_{<t}) \;\middle|\; v_{n_i} \in N \right\}$
    \STATE \textit{$P_t$}: $\left\{f_\theta(\cdot \mid v_{p_i}, x, y_{<t}) \;\middle|\; v_{p_i} \in P \right\}$
    \STATE \textit{sum}($\cdot$): The element-wise sum of all logits.
    \STATE \textbf{Compute $\textit{RVCD}_{y_t}$:} 
    \[
    \hspace{-2.4em} 
    \begin{aligned}
    \textit{RVCD}_{logit_t} 
    & = \\
    & \!\!\!\!\!\!\!\!\!\!\!\!\!\!\!\!\!\!\!\!\!\!\!\!\!\!\!\big(1 + \alpha \cdot \textit{len}(N) - \beta \cdot \textit{len}(P)\big) \\
    & \!\!\!\!\!\!\!\!\!\!\!\!\!\!\!\!\!\!\!\!\!\!\!\!\!\!\!\!\cdot f_\theta(\cdot \mid v, x, y_{<t}) & \\ 
    & \!\!\!\!\!\!\!\!\!\!\!\!\!\!\!\!\!\!\!\!\!\!\!\!\!\!\!\!- \big(\alpha \cdot \textit{sum}(N_t) - \beta \cdot \textit{sum}(P_t)\big).
    \end{aligned}
    \]
    \[
    \hspace{0.9em} 
    \textit{RVCD}_{y_t} = {\small\arg\max( \softmax[\textit{RVCD}_{logit_t}])}
    \]
\UNTIL{Obtain decoded sequence $RVCDy_0$, \dots, $RVCDy_t$.}
\STATE Combine $RVCDy_0$, \dots, $RVCDy_t$ into \textit{RVCD output text}.
\end{algorithmic}
\end{algorithm} 
These experiments comprehensively evaluate the accuracy and quality of the text generated by RVCD from the perspective of OH mitigation.\\[0.5em]
\textbf{Baselines.}
Given that RVCD is a decoding method, we compared it with general decoding strategies such as greedy decoding and beam search, as well as established state-of-the-art (SOTA) decoding methods:
DOLA~\cite{chuang2023dola},
OPERA~\cite{huang2023opera},
VCD~\cite{leng2023visual}, and
HALC~\cite{chen2024halc}.
All evaluations were conducted under identical experimental settings.\\[0.5em]
\textbf{LVLM Backbones.}
We adopted three widely used 7B backbones for our experiments:
MiniGPT-4 V2 with vicuna-7b~\cite{chen2023minigptv2},
LLaVA-1.5~\cite{liu2023visual}, and
mPLUG-Owl2~\cite{ye2023mplugowl2}.
These backbones were selected to enable the most direct comparison with results from prior decoding-based studies~\cite{chen2024halc, zhuang2024game}. 
\subsection{CHAIR, BLEU and POPE on MSCOCO}
To reproduce the evaluation methodologies of prior studies and assess the performance of RVCD under the same conditions, we used their identical experimental setup~\cite{huang2023opera, liu2023visual}. Specifically, we employed the validation set of MSCOCO 2014~\cite{lin2014mscoco}, randomly sampling 500 images five times with replacement and reporting the mean and standard deviation for CHAIR and POPE metrics.

\noindent \textbf{CHAIR.}
CHAIR (Caption Hallucination Assessment with Image Relevance) quantifies OH in image captioning tasks~\cite{rohrbach2018object}. It assumes the existence of a ground truth object list for each image and uses a dictionary to map objects in the output captions to representative MSCOCO synonyms. The primary metrics are:\\
CHAIR$_I$: The ratio of objects in the captions that do not appear in the ground truth object list to the total number of objects mentioned in the captions.
CHAIR$_S$: The proportion of sentences with hallucination (i.e., sentences where CHAIR$_I$ is non-zero).
Lower values of CHAIR$_I$ and CHAIR$_S$ indicate lower levels of OH. In line with prior studies, we performed image captioning using the same prompt:
“Please describe this image in detail.”
The results are shown in Table~\ref{tab:chair_table}.
In addition to CHAIR$_I$ and CHAIR$_S$, we provide BLEU scores~\cite{papineni2002bleu}. 
\begin{table*}[t]
\centering
\small
\setlength{\tabcolsep}{2.5pt} 
\renewcommand{\arraystretch}{1.2} 
\resizebox{\linewidth}{!}{ 
\begin{tabularx}{\textwidth}{l|ccc|ccc|ccc}
\toprule
\multirow{2}{*}{\raisebox{-0.2em}{\textbf{Methods}}} & \multicolumn{3}{c|}{\textbf{LLaVA-1.5}} & \multicolumn{3}{c|}{\textbf{MiniGPT-4}} & \multicolumn{3}{c}{\textbf{mPLUG-Owl2}} \\
\cline{2-4} \cline{5-7} \cline{8-10}
 & CHAIR$_S$ ↓ & CHAIR$_I$ ↓ & BLEU ↑ & CHAIR$_S$ ↓ & CHAIR$_I$ ↓ & BLEU ↑ & CHAIR$_S$ ↓ & CHAIR$_I$ ↓ & BLEU ↑ \\
\midrule
Greedy         & 22.08\scriptsize{$\pm$1.05} & 7.08\scriptsize{$\pm$0.37} & 16.06\scriptsize{$\pm$0.17} & 20.32\scriptsize{$\pm$1.45} & 7.03\scriptsize{$\pm$0.59} & 16.17\scriptsize{$\pm$0.26} & 23.87\scriptsize{$\pm$0.92} & 8.77\scriptsize{$\pm$0.41} & 15.43\scriptsize{$\pm$0.20} \\
Beam Search    & 20.60\scriptsize{$\pm$1.39} & 6.95\scriptsize{$\pm$0.28} & 16.33\scriptsize{$\pm$0.09} & 20.64\scriptsize{$\pm$0.74} & 7.32\scriptsize{$\pm$0.63} & 16.55\scriptsize{$\pm$0.26} & 21.60\scriptsize{$\pm$1.14} & 8.02\scriptsize{$\pm$0.39} & 15.61\scriptsize{$\pm$0.28} \\
DoLA           & 21.36\scriptsize{$\pm$0.65} & 6.82\scriptsize{$\pm$0.20} & 16.11\scriptsize{$\pm$0.12} & 20.36\scriptsize{$\pm$1.87} & 7.08\scriptsize{$\pm$0.68} & 16.10\scriptsize{$\pm$0.25} & 24.40\scriptsize{$\pm$1.65} & 8.76\scriptsize{$\pm$0.52} & 15.46\scriptsize{$\pm$0.24} \\
OPERA          & 18.72\scriptsize{$\pm$1.20} & 6.56\scriptsize{$\pm$0.39} & \textbf{16.65}\scriptsize{$\pm$0.21} & 19.44\scriptsize{$\pm$1.71} & 7.22\scriptsize{$\pm$0.71} & 17.77\scriptsize{$\pm$0.25} & 20.24\scriptsize{$\pm$0.79} & 7.80\scriptsize{$\pm$0.38} & 15.49\scriptsize{$\pm$0.10} \\
VCD            & 23.24\scriptsize{$\pm$1.17} & 7.73\scriptsize{$\pm$0.28} & 14.97\scriptsize{$\pm$0.24} & 21.72\scriptsize{$\pm$1.26} & 8.08\scriptsize{$\pm$0.40} & 15.92\scriptsize{$\pm$0.24} & 26.72\scriptsize{$\pm$1.57} & 10.08\scriptsize{$\pm$0.60} & 14.27\scriptsize{$\pm$0.29} \\
HALC           & 18.60\scriptsize{$\pm$0.70} & 6.03\scriptsize{$\pm$0.32} & 16.32\scriptsize{$\pm$0.13} & 15.36\scriptsize{$\pm$2.26} & 5.55\scriptsize{$\pm$0.71} & \textbf{17.83}\scriptsize{$\pm$0.38} & 21.08\scriptsize{$\pm$1.37} & 7.54\scriptsize{$\pm$0.49} & \textbf{15.63}\scriptsize{$\pm$0.26} \\
\midrule
\textbf{RVCD}  & \boldmath\textbf{11.32}\scriptsize{$\pm$0.92} & \textbf{3.87}\scriptsize{$\pm$0.35} & 15.48\scriptsize{$\pm$0.13} & \textbf{9.00}\scriptsize{$\pm$1.17} & \textbf{3.61}\scriptsize{$\pm$0.50} & 15.98\scriptsize{$\pm$0.33} & \textbf{10.04}\scriptsize{$\pm$1.77} & \textbf{3.73}\scriptsize{$\pm$0.54} & 14.78\scriptsize{$\pm$0.22} \\
\bottomrule
\end{tabularx}
} 

\caption{The averages and sample standard deviations of CHAIR and BLEU metrics with different decoding baselines were calculated over five different sampling seeds, each involving a random sampling of 500 instances from the MSCOCO dataset. Lower scores in CHAIR$_S$ and CHAIR$_I$ indicate less OH, while higher BLEU scores reflect better caption quality.}
\label{tab:chair_table}
\end{table*}
\begin{table*}[t]
\centering
\small
\setlength{\tabcolsep}{2.25pt} 
\renewcommand{\arraystretch}{1.2} 
\begin{tabularx}{\textwidth}{l|ccc|ccc|ccc}
\toprule
\textbf{Methods} & \multicolumn{3}{c|}{\textbf{LLaVA-1.5}} & \multicolumn{3}{c|}{\textbf{MiniGPT-4}} & \multicolumn{3}{c}{\textbf{mPLUG-Owl2}} \\
\cline{2-4} \cline{5-7} \cline{8-10}
 & Accuracy ↑ & Precision ↑ & $F_{1}$ ↑ & Accuracy ↑ & Precision ↑ & $F_{1}$ ↑ & Accuracy ↑ & Precision ↑ & $F_{1}$ ↑ \\
\midrule
Greedy         & 72.19\scriptsize{$\pm$6.10} & 65.28\scriptsize{$\pm$5.49} & 77.86\scriptsize{$\pm$3.88} & 62.98\scriptsize{$\pm$9.36} & 58.72\scriptsize{$\pm$7.28} & 72.24\scriptsize{$\pm$5.38} & 74.36\scriptsize{$\pm$5.89} & 67.23\scriptsize{$\pm$5.59} & 79.23\scriptsize{$\pm$3.86} \\
Beam Search    & 78.27\scriptsize{$\pm$4.47} & 71.94\scriptsize{$\pm$4.96} & 81.28\scriptsize{$\pm$3.17} & 67.51\scriptsize{$\pm$7.49} & 62.67\scriptsize{$\pm$6.89} & 73.82\scriptsize{$\pm$4.68} & 80.17\scriptsize{$\pm$5.08} & 74.30\scriptsize{$\pm$6.01} & 82.64\scriptsize{$\pm$3.72} \\
DoLA           & 72.48\scriptsize{$\pm$6.10} & 65.54\scriptsize{$\pm$5.54} & 78.04\scriptsize{$\pm$3.90} & 72.26\scriptsize{$\pm$3.96} & 75.41\scriptsize{$\pm$7.04} & 70.86\scriptsize{$\pm$3.09} & 74.56\scriptsize{$\pm$5.85} & 67.43\scriptsize{$\pm$5.59} & 79.34\scriptsize{$\pm$3.85} \\
OPERA          & 74.43\scriptsize{$\pm$4.11} & 67.43\scriptsize{$\pm$3.93} & 78.94\scriptsize{$\pm$2.71} & 67.49\scriptsize{$\pm$6.74} & 62.41\scriptsize{$\pm$6.05} & 73.89\scriptsize{$\pm$4.16} & 79.01\scriptsize{$\pm$5.62} & 72.71\scriptsize{$\pm$6.36} & 81.98\scriptsize{$\pm$4.03} \\
VCD            & 69.86\scriptsize{$\pm$3.48} & 63.44\scriptsize{$\pm$3.02} & 75.87\scriptsize{$\pm$2.14} & 61.79\scriptsize{$\pm$3.39} & 58.97\scriptsize{$\pm$3.07} & 67.34\scriptsize{$\pm$2.02} & 73.66\scriptsize{$\pm$3.80} & 67.33\scriptsize{$\pm$3.76} & 77.93\scriptsize{$\pm$2.53} \\
HALC           & 72.48\scriptsize{$\pm$6.10} & 65.54\scriptsize{$\pm$5.54} & 78.04\scriptsize{$\pm$3.90} & 72.26\scriptsize{$\pm$3.96} & 75.41\scriptsize{$\pm$7.04} & 70.86\scriptsize{$\pm$3.09} & 74.54\scriptsize{$\pm$5.85} & 67.42\scriptsize{$\pm$5.59} & 79.33\scriptsize{$\pm$3.84} \\
\midrule
\textbf{RVCD}  & \textbf{88.54\scriptsize{$\pm$2.59}} & \textbf{89.92}\scriptsize{$\pm$4.70} & \textbf{88.43}\scriptsize{$\pm$2.33} & \textbf{85.96}\scriptsize{$\pm$2.37} & \textbf{88.14}\scriptsize{$\pm$4.34} & \textbf{85.63}\scriptsize{$\pm$2.05} & \textbf{87.45}\scriptsize{$\pm$1.64} & \textbf{87.91}\scriptsize{$\pm$2.89} & \textbf{87.41}\scriptsize{$\pm$1.44} \\
\bottomrule
\end{tabularx}
\captionsetup{aboveskip=5pt, belowskip=-12.5pt}
\caption{POPE evaluation results on MSCOCO dataset of LVLMs with different decoding baselines designed to mitigate OH. Higher accuracy, precision, and F$_1$ indicate better performance.}

\label{tab:pope-table}
\end{table*}
\begin{figure}[H] 
    \centering
    \captionsetup{aboveskip=5pt, belowskip=-15pt}
    \includegraphics[width=\linewidth]{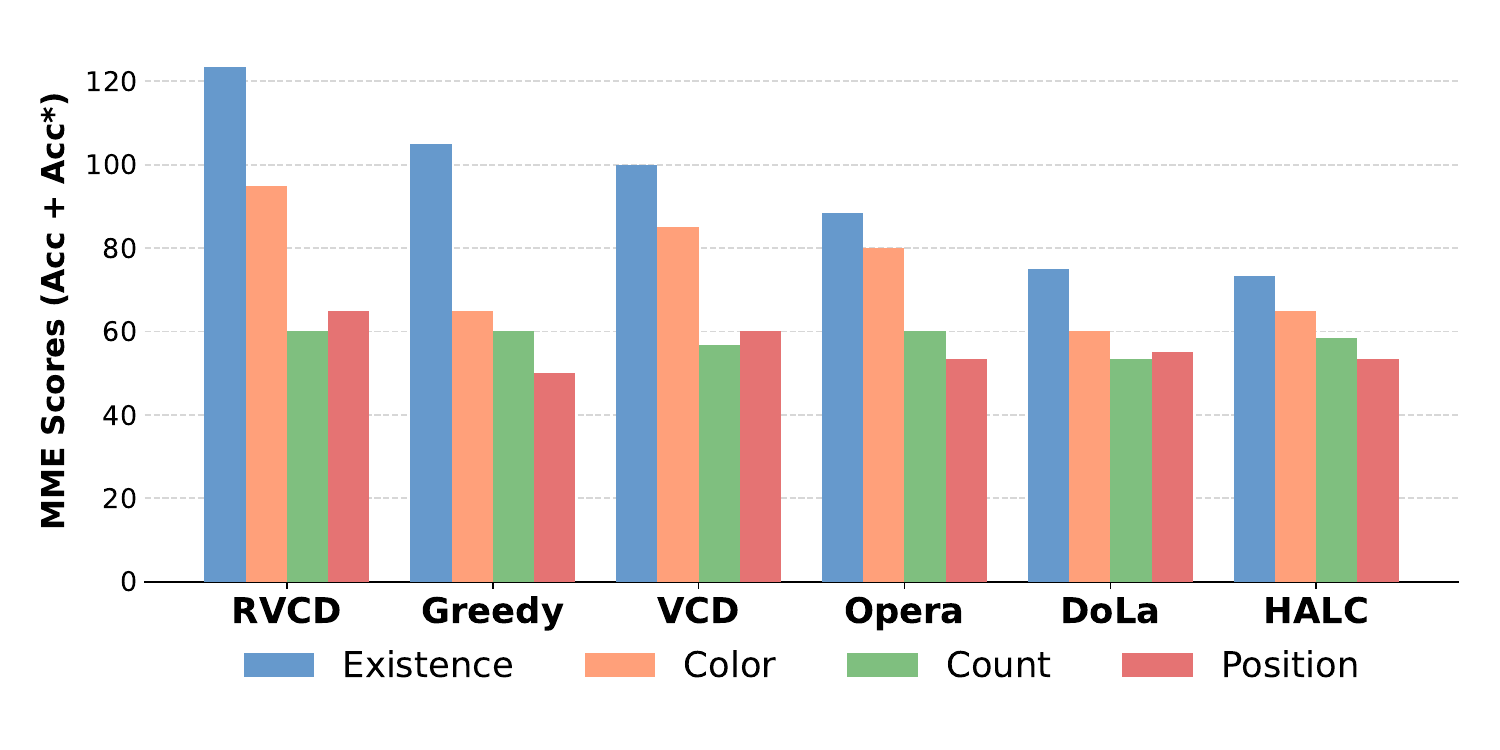} 
    \caption{Comparison of different decoding baselines on MME Metric with llava-1.5 as a backbone LVLM. Refer to Table~\ref{tab:mme_details} for detailed information, including MiniGPT-4 and mPLUG-Owl2.}
    \label{fig:mme}
\end{figure}


\begin{table*}[t]
\centering
\small
\captionsetup{aboveskip=5pt, belowskip=-10pt}
\setlength{\tabcolsep}{1.5pt} 
\renewcommand{\arraystretch}{1.2} 
\resizebox{\linewidth}{!}{ 
\begin{tabularx}{\textwidth}{l|ccc|ccc|ccc}
\toprule
\multirow{2}{*}{\raisebox{-0.2em}{\textbf{$N$, $P$ Settings}}} & \multicolumn{3}{c|}{\textbf{LLaVA-1.5}} & \multicolumn{3}{c|}{\textbf{MiniGPT-4}} & \multicolumn{3}{c}{\textbf{mPLUG-Owl2}} \\
\cline{2-4} \cline{5-7} \cline{8-10}
 
 & CHAIR$_S$ ↓ & CHAIR$_I$ ↓ & BLEU ↑ &  CHAIR$_S$ ↓ & CHAIR$_I$ ↓ & BLEU ↑ & CHAIR$_S$ ↓ & CHAIR$_I$ ↓ & BLEU ↑ \\
\midrule

gt ($ann$), hal ($ann$) & 30.2\scriptsize{$\pm$0.86} & 10.75\scriptsize{$\pm$0.36} &  12.06\scriptsize{$\pm$0.19} & 14.20\scriptsize{$\pm$2.44} & 9.08\scriptsize{$\pm$1.15} & 12.88\scriptsize{$\pm$0.39} & 27.88\scriptsize{$\pm$1.54} & 11.76\scriptsize{$\pm$0.69} & 11.88\scriptsize{$\pm$0.24} \\ 
gt+hal ($ann$), $\emptyset$ & 29.04\scriptsize{$\pm$0.72} & 10.05\scriptsize{$\pm$0.26} &  11.78\scriptsize{$\pm$0.21} & 12.96\scriptsize{$\pm$1.58} & 7.87\scriptsize{$\pm$0.67} & 12.51\scriptsize{$\pm$0.30} & 25.52\scriptsize{$\pm$1.82} & 10.33\scriptsize{$\pm$1.02} & 11.60\scriptsize{$\pm$0.29} \\ 
hal ($yl\ 3$), gt ($yl\ 3$) & 12.84\scriptsize{$\pm$1.11} & 4.48\scriptsize{$\pm$0.65} &  14.99\scriptsize{$\pm$0.21} & 9.28\scriptsize{$\pm$0.99} & 3.68\scriptsize{$\pm$0.39} & 15.81\scriptsize{$\pm$0.24} & 12.12\scriptsize{$\pm$2.38} & 4.58\scriptsize{$\pm$0.67} & 14.45\scriptsize{$\pm$0.22} \\ 
hal ($yl\ 8$), gt ($yl\ 8$) & 12.4\scriptsize{$\pm$0.70} & 3.98\scriptsize{$\pm$0.34} &  15.18\scriptsize{$\pm$0.10} & 8.84\scriptsize{$\pm$0.50} & 3.47\scriptsize{$\pm$0.17} & 15.99\scriptsize{$\pm$0.37} & 10.68\scriptsize{$\pm$2.27} & 3.96\scriptsize{$\pm$0.80} & 14.74\scriptsize{$\pm$0.24} \\ 
\textbf{hal (\textbf{ann}), gt (\textbf{ann})} & \textbf{8.44}\scriptsize{$\pm$0.84} & \textbf{2.77}\scriptsize{$\pm$0.12} &  \textbf{15.61}\scriptsize{$\pm$0.13} & \textbf{6.48}\scriptsize{$\pm$0.36} & \textbf{2.48}\scriptsize{$\pm$0.13} & \textbf{16.10}\scriptsize{$\pm$0.28} & \textbf{7.96}\scriptsize{$\pm$1.00} & \textbf{2.84}\scriptsize{$\pm$0.46} & \textbf{15.02}\scriptsize{$\pm$0.31} \\ 
\midrule
\end{tabularx}
} 
\caption{Performance under CHAIR and BLEU with increasing detection rates, simulated using annotations or Object Detection models. Detection rate in order: 0\%, treating all objects as N, detection with YOLOv3 ($yl\ 3$), detection with YOLOv8x ($yl\ 8$), 100\%. $\alpha$, $\beta$ were set to 1 and 0, respectively for this experiment.}
\label{tab:NP}
\end{table*}

\noindent \textbf{BLEU.} 
BLEU (Bilingual Evaluation Understudy)~\cite{papineni2002bleu} is a traditional metric for evaluating caption quality based on n-gram matching rates.
As shown in Table~\ref{tab:chair_table}, our RVCD significantly reduces OH compared to other decoding methods while maintaining comparable BLEU scores. This demonstrates the effectiveness of retrieval-based negative and positive logits adjustment of our RVCD.

\noindent \textbf{POPE.}
POPE (Polling-based Object Probing Evaluation)~\cite{li2023evaluating} evaluates OH through binary classification. The LVLM is prompted to answer "yes" or "no" to whether specific objects exist in an image. POPE provides three evaluation scenarios:
random, popular, and adversarial.
Detailed descriptions of these options are provided by~\citeauthorandyear{li2023evaluating}. As shown in Table~\ref{tab:pope-table}, our RVCD significantly outperformed all other methods in accuracy, precision, and F$_1$ scores. This demonstrates that RVCD effectively suppresses OH of LVLMs.

\noindent\textbf{MME.}
MME (The Multimodal Large Language Model Evaluation)~\cite{fu2023mme} is a quantitative evaluation benchmark, similar to CHAIR, BLEU, and POPE, but provides diverse subsets for evaluation. Following prior studies~\cite{yin2023woodpecker, leng2023visual, chen2024halc, zhuang2024game}, we evaluated RVCD on the "existence", "count", "position", and "color" subsets to comprehensively assess OH. Notably, the existence subset highlights RVCD's ability to effectively transfer the object detection capabilities of OD models to LVLMs.

\noindent\textbf{LLaVA-Bench.}
LLaVA-Bench~\cite{liu2023visual2} consists of 24 images, each paired with highly accurate and detailed human-generated descriptions. 
\begin{figure*}[ht]
    \captionsetup{aboveskip=5pt, belowskip=-10pt}
    \centering
    \includegraphics[width=\textwidth]{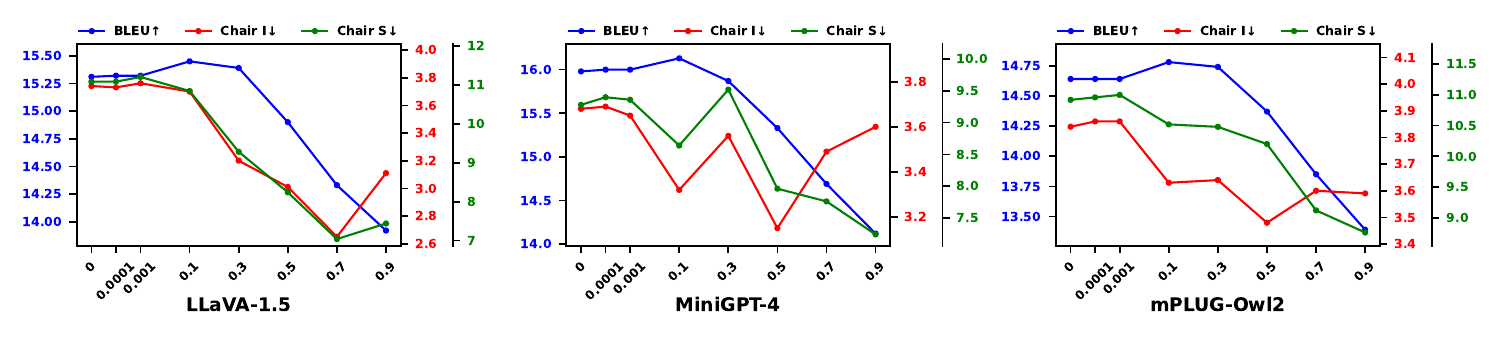}
    \caption{
    BLEU and CHAIR scores based on the variation of the $\beta$ value, with $\alpha$ set to 1. The mean of five samples, each consisting of 500 instances, sampled with replacement from the MSCOCO 2014 validation dataset.}
    \label{fig:vis_Beta ablation}
\end{figure*}
This benchmark includes three types of questions:
simple question answering (conversation),
detailed descriptions,
and complex reasoning. To qualitatively evaluate RVCD, we leveraged LLaVA-Bench as a case study, following the methodologies of prior studies. The detailed results are provided in Appendix~\ref{sec:appendix_llavabench}.

\section{Analysis and Ablation Studies}
\subsection{Effect of Accurate Detection}
In Table~\ref{tab:NP}, the N and P settings, gt ($ann$), hal ($ann$) indicate a 0\% detection rate when based on annotations. gt+hal ($ann$), $\emptyset$ assumes that all objects in the draft caption are treated as negative logits. hal ($yl\ 3, 8$), gt ($yl\ 3, 8$) refer to the application of object detection models YOLOv3~\cite{redmon2018yolov3} and YOLOv8x~\cite{yolov8} with confidence threshold 0.25 (default setting), and hal ($ann$), gt ($ann$) represent a scenario where the detection rate is 100\% based on annotations. For a fair comparison between Setting gt+hal ($ann$), $\emptyset$ and other settings, $\alpha$, $\beta$ were set to 1 and 0, respectively for this experiment. CHAIR and BLEU scores for captioning improve with increased detection accuracy. This shows that our RVCD becomes increasingly effective as detection accuracy improves, suggesting that advancements in object detection models can have a direct positive impact on RVCD performance.

\subsection{Ablation Study of $\alpha$ and $\beta$}
The statistical details of the ablation of $\alpha$ and $\beta$ are presented in Appendix~\ref{sec:alpha_beta}. Increasing $\alpha$ strengthens the regulation of hallucinated (Hal) objects, benefiting the CHAIR score. However, as observed in Section~\ref{tab:co_occur_section}, it also unintentionally removes parts of the ground truth (GT) objects. The CHAIR gains from Hal object removal outweigh the CHAIR losses caused by GT object degradation, ultimately resulting in a net benefit as $\alpha$ increases. Nevertheless, the gradual decline in BLEU indicates that the issue of GT object degradation persists (Table~\ref{tab:alpha_ablation}).

To address this, we introduced $\beta$ and positive logits at a level that restores the degraded GT objects, as represented in Equation~\ref{tab:main_equation}. $\alpha$ maximizes the benefits of Hal object regulation at $\alpha$ = 1, while $\beta$ maximizes GT recovery at $\beta$ = 0.1. As shown in Figure~\ref{fig:vis_Beta ablation}, at $\beta$ = 0.1, the recovery of the degraded GT objects leads to gains across CHAIR$_S$, CHAIR$_I$, and BLEU. Therefore, we propose ($\alpha$, $\beta$) = (1, 0.1) as the optimal setting of RVCD.

\subsection{Decoding Latency Analysis on Image Captioning}

Different state-of-the-art (SOTA) decoding methods vary in the frequency and timing of external model calls, depending on their underlying methodologies. Furthermore, the computational cost can differ across specific decoding steps.
As such, empirically measuring token generation latency is a reasonable and appropriate approach.

To this end, we evaluate decoding efficiency by sampling 500 images with replacement from the MSCOCO 2014 validation set, repeating the process five times and reporting the mean and standard deviation across runs. For details on the decoding configurations, refer to Appendix~\ref{sec:appendixG}.

RVCD operates by removing hallucinated objects \( N \) from the greedy decoded caption and reintroducing ground-truth objects \( P \). As demonstrated in our ablation study (Table~\ref{tab:NP}, 4th row), even a lightweight adjustment—removing only the hallucinated objects \( N \) without reintroducing ground-truth objects \( P \) (i.e., \( \beta = 0 \))—achieves superior OH reduction compared to other state-of-the-art decoding methods. Introducing \( P \) (\( \beta = 0.1 \)), as in the full RVCD pipeline, can further enhance performance (Figure~\ref{fig:vis_Beta ablation}).

RVCD exhibits significantly lower decoding latency than previous state-of-the-art methods such as HALC~\cite{chen2024halc} and OPERA~\cite{huang2023opera}. Despite incorporating the generation of a greedy decoded draft, comparison with objects detected via YOLO, and contrastive decoding processes involving multiple explicit images, RVCD achieves superior performance in both output quality and decoding efficiency.

\begin{table}[h]
\centering
\small
\begin{tabular}{lcc}
\toprule
\textbf{Methods} & \textbf{Avg. Latency (s/token)} & \textbf{Relative} \\
\midrule
Greedy     & 0.034 ± \scriptsize{0.002} & $1.000\times$ \\
DoLa       & 0.048 ± \scriptsize{0.001} & $1.416\times$ \\
VCD        & 0.073 ± \scriptsize{0.002} & $2.174\times$ \\
OPERA      & 0.341 ± \scriptsize{0.004} & $10.128\times$ \\
HALC       & 0.800 ± \scriptsize{0.017} & $23.795\times$ \\
\midrule
\textbf{RVCD ($\boldsymbol{\beta = 0}$)} & 0.143 ± \scriptsize{0.003} & $4.242\times$ \\
\textbf{RVCD ($\boldsymbol{\beta \ne 0}$)}  & 0.204 ± \scriptsize{0.004} & $6.053\times$ \\
\bottomrule
\end{tabular}
\caption{Decoding Latency per Token.}
\end{table}


\section{Conclusion}
We propose RVCD, an advanced train-free decoding-based plug-and-play method that significantly alleviates the Object Hallucination (OH) problem of LVLMs. Inspired by prior studies leveraging the idea of Visual Contrastive Decoding (VCD), RVCD maximizes the potential of VCD through negative and positive logits generated from explicit images retrieved from a single-concept image database. Comprehensive experiments demonstrate that RVCD significantly outperforms other state-of-the-art (SOTA) decoding methods.

\section{Limitations}
RVCD adjusts the final output by leveraging additional information from multiple negative and positive logits generated from an image that explicitly represents a single concept. This approach requires generating logits for the number of words in the evaluation dictionary that match the set of objects mentioned in the greedy decoded draft caption, excluding duplicates, at each decoding step. Consequently, if the draft caption mentions an overly diverse and large number of words, it may result in a disadvantage in token latency. In future work, we will focus on constructing a more efficient reference image database to mitigate this issue.

\section*{Ethics Statement}
The purpose of this study is to introduce a decoding method to mitigate Object Hallucination (OH) in Large Vision-Language Models (LVLMs). In this process, we utilized transformer-based LVLMs and a diffusion-based image generation model. While these models, as generative AI, may produce uncontrollable or disloyal outputs. Nevertheless, the goal of RVCD is to mitigate such disloyal outputs and OH, aligning with ethical review standards. Additionally, all our experiments were conducted using public datasets, and every image in the AI-generated image database we provide was manually checked and contains general object representations that do not pose ethical concerns. 

\section{Acknowledgements}
This research was supported by the Culture, Sports
and Tourism R\&D Program through the Korea
Creative Content Agency grant funded by the
Ministry of Culture, Sports and Tourism in 2024
(Project Name: Developing a generative AI story
platform for Fanfiction, Project Number: RS-2024-
00442270). This research was partly supported by
an IITP grant funded by the Korean Government
(MSIT) (No. RS-2020-II201361, Artificial Intel-
ligence Graduate School Program (Yonsei Univer-
sity)).


\clearpage

\appendix

\section{Data License and Usage}
\label{sec:appendix}

Our experiment was conducted using the MSCOCO validation 2014 dataset~\cite{lin2014mscoco} for CHAIR, BLEU, and POPE. Additionally, we evaluated the quality of image captioning both quantitatively and qualitatively through MME~\cite{fu2023mme} and LLAVA-bench~\cite{liu2023visual2}. We clarify that all of these datasets are publicly available for research purposes and were utilized to assess the image captioning performance of various decoding-based methods. Additionally, our generated single concept AI images were created through the image generation model FLUX.1-dev, which is under Non-Commercial License~\cite{FLUX1_LICENSE}.

\section{Computational Resources}
\label{sec:appendix}
We adopted widely used 7B-sized LVLM backbones in our experimental environment, including MiniGPT-4 V2 with Vicuna-7B~\cite{chen2023minigptv2}, LLaVA-1.5~\cite{liu2023visual}, and mPLUG-Owl2~\cite{ye2023mplugowl2}. All experiments were conducted without any training, performing only inference, and were executed on a single NVIDIA A100 GPU.

\section{Detection Precision on Draft Captions}
\label{appendix_precision}
As shown in Figure~\ref{tab:detector_bar}, the precision of each LVLM when answering a VQA question with "yes" or "no" about the presence of an object mentioned in its generated caption is expressed as GT/Hal (LVLM). Similarly, when treating objects detected/not detected by YOLO as equivalent to the LVLM's "yes" or "no" answers, YOLO's precision is represented as GT/Hal (YOLO).
Hal (LVLM) is noticeably worse, indicating that errors occurred in most cases where LVLMs gave negative answers in the VQA task. In other words, there were many False Negatives where the objects actually existed, but the LVLMs stated that they did not. This suggests that the object detection capability of LVLMs is insufficient to self-correct hallucinated objects in the draft captions.

Table~\ref{tab:detect_ablation} represents the statistical details of Figure~\ref{tab:detector_bar}.

\begin{table}[h]
\centering
\small
\caption{Statistical details of detection precision on draft captions. The mean and standard deviation of five samples, each consisting of 500 instances, sampled with replacement from the MSCOCO 2014 validation dataset.}
\captionsetup{aboveskip=5pt, belowskip=-10pt}
\setlength{\tabcolsep}{3pt} 
\renewcommand{\arraystretch}{1.2} 
\begin{tabular}{l|c|c|c} 
\toprule
\textbf{Precision} & \textbf{LLaVA-1.5} & \textbf{MiniGPT-4} & \textbf{mPLUG-Owl2} \\
\midrule
Hal (YOLO) & 90.05\scriptsize{$\pm$0.85} & 91.75\scriptsize{$\pm$4.39} & 91.23\scriptsize{$\pm$2.95} \\ 
Hal (LVLM) & 28.77\scriptsize{$\pm$3.01} & 57.49\scriptsize{$\pm$5.24} & 18.10\scriptsize{$\pm$2.57} \\ 
GT (YOLO) & 91.12\scriptsize{$\pm$0.28} & 91.23\scriptsize{$\pm$0.78} & 91.94\scriptsize{$\pm$0.47} \\ 
GT (LVLM) & 98.21\scriptsize{$\pm$0.38} & 89.08\scriptsize{$\pm$1.35} & 99.05\scriptsize{$\pm$0.37} \\ 
\midrule
\end{tabular}
\label{tab:detect_ablation}
\end{table}

\section{Comprehensive POPE Results}
\label{sec:appendix}
In the POPE evaluation, to create a fair environment similar to previous studies~\cite{chen2024halc, zhuang2024game}, we combined the entire query of POPE with an initially greedy decoded answer (yes/no) and used it as a draft caption for RVCD. Accordingly, the detector determines whether the object mentioned in the draft caption actually exists in the image and conveys this judgment to the LVLMs. This is similar to how HALC~\cite{chen2024halc} combines the entire query with an initial answer (yes/no) to form a text prompt, allowing the detection model to provide grounding for the focal area of the query in their POPE evaluation. The statistical details are presented in Table~\ref{tab:pope_all_1} and Table~\ref{tab:pope_all_2}.

\begin{table*}[t]
    \centering
    \captionsetup{aboveskip=5pt, belowskip=-10pt}
    \caption{Comparison of the mean of five POPE results on MSCOCO dataset with different decoding baselines under the ‘random’ and ‘popular’ settings. Higher accuracy (Acc.), precision (Prec.), and F$_1$ score indicate better performance.}
    \label{tab:pope_all_1}
    \resizebox{\textwidth}{!}{
    \begin{tabularx}{\textwidth}{@{}p{2cm}p{2.5cm}p{2cm}XXXX@{}}
    \toprule
    Setting & Model & Decoding & Accuracy & Precision & Recall & $F_{1}$ Score \\
    \midrule
    \multirow{21}{*}{Random} 
    & \multirow{7}{*}{LLaVA-1.5} 
    & Greedy & 79.93\scriptsize{$\pm$0.74} & 72.37\scriptsize{$\pm$0.74} & 96.82\scriptsize{$\pm$0.37} & 82.83\scriptsize{$\pm$0.55} \\
    & & Beam Search & 83.65\scriptsize{$\pm$0.29} & 78.05\scriptsize{$\pm$0.36} & 93.64\scriptsize{$\pm$0.52} & 85.13\scriptsize{$\pm$0.26} \\
    & & DoLa & 80.22\scriptsize{$\pm$0.76} & 72.69\scriptsize{$\pm$0.77} & 96.81\scriptsize{$\pm$0.34} & 83.03\scriptsize{$\pm$0.57} \\
    & & OPERA & 79.35\scriptsize{$\pm$0.88} & 72.24\scriptsize{$\pm$0.86} & 95.36\scriptsize{$\pm$0.54} & 82.20\scriptsize{$\pm$0.68} \\
    & & VCD & 74.28\scriptsize{$\pm$0.41} & 67.30\scriptsize{$\pm$0.35} & 94.45\scriptsize{$\pm$0.21} & 78.60\scriptsize{$\pm$0.30} \\
    & & HALC & 80.22\scriptsize{$\pm$0.76} & 72.69\scriptsize{$\pm$0.77} & 96.81\scriptsize{$\pm$0.34} & 83.03\scriptsize{$\pm$0.57} \\
    & & RVCD & 91.33\scriptsize{$\pm$0.34} & 95.16\scriptsize{$\pm$0.48} & 87.09\scriptsize{$\pm$0.82} & 90.94\scriptsize{$\pm$0.39} \\
    \cmidrule{2-7}
    & \multirow{7}{*}{MiniGPT-4} 
    & Greedy & 75.70\scriptsize{$\pm$0.41} & 68.63\scriptsize{$\pm$0.44} & 94.68\scriptsize{$\pm$0.37} & 79.58\scriptsize{$\pm$0.27} \\
    & & Beam Search & 77.58\scriptsize{$\pm$0.92} & 71.97\scriptsize{$\pm$1.06} & 90.40\scriptsize{$\pm$0.31} & 80.13\scriptsize{$\pm$0.65} \\
    & & DoLa & 77.55\scriptsize{$\pm$0.65} & 84.84\scriptsize{$\pm$0.96} & 67.10\scriptsize{$\pm$1.58} & 74.92\scriptsize{$\pm$0.92} \\
    & & OPERA & 76.49\scriptsize{$\pm$0.69} & 70.53\scriptsize{$\pm$0.80} & 91.02\scriptsize{$\pm$0.41} & 79.47\scriptsize{$\pm$0.46} \\
    & & VCD & 66.25\scriptsize{$\pm$0.76} & 63.03\scriptsize{$\pm$0.70} & 78.61\scriptsize{$\pm$0.90} & 69.96\scriptsize{$\pm$0.62} \\
    & & HALC & 77.55\scriptsize{$\pm$0.65} & 84.84\scriptsize{$\pm$0.96} & 67.10\scriptsize{$\pm$1.58} & 74.92\scriptsize{$\pm$0.92} \\
    & & RVCD & 87.96\scriptsize{$\pm$1.01} & 91.84\scriptsize{$\pm$2.25} & 83.37\scriptsize{$\pm$0.44} & 87.38\scriptsize{$\pm$0.91} \\
    \cmidrule{2-7}
    & \multirow{7}{*}{mPLUG-Owl2} 
    & Greedy & 81.68\scriptsize{$\pm$0.95} & 74.33\scriptsize{$\pm$0.99} & 96.80\scriptsize{$\pm$0.37} & 84.08\scriptsize{$\pm$0.72} \\
    & & Beam Search & 86.22\scriptsize{$\pm$0.47} & 81.65\scriptsize{$\pm$0.61} & 93.42\scriptsize{$\pm$0.25} & 87.14\scriptsize{$\pm$0.40} \\
    & & DoLa & 81.82\scriptsize{$\pm$0.87} & 74.53\scriptsize{$\pm$0.94} & 96.72\scriptsize{$\pm$0.32} & 84.18\scriptsize{$\pm$0.66} \\
    & & OPERA & 85.88\scriptsize{$\pm$0.64} & 80.70\scriptsize{$\pm$0.75} & 94.34\scriptsize{$\pm$0.52} & 86.99\scriptsize{$\pm$0.56} \\
    & & VCD & 78.45\scriptsize{$\pm$0.61} & 72.10\scriptsize{$\pm$0.71} & 92.80\scriptsize{$\pm$0.51} & 81.15\scriptsize{$\pm$0.44} \\
    & & HALC & 81.81\scriptsize{$\pm$0.87} & 74.51\scriptsize{$\pm$0.95} & 96.72\scriptsize{$\pm$0.28} & 84.17\scriptsize{$\pm$0.65} \\
    & & RVCD & 89.05\scriptsize{$\pm$0.48} & 90.75\scriptsize{$\pm$0.83} & 86.97\scriptsize{$\pm$0.26} & 88.82\scriptsize{$\pm$0.45} \\
    
    \midrule
    
    \multirow{21}{*}{Popular} 
    & \multirow{7}{*}{LLaVA-1.5} 
    & Greedy & 70.72\scriptsize{$\pm$1.47} & 63.63\scriptsize{$\pm$1.20} & 96.82\scriptsize{$\pm$0.37} & 76.78\scriptsize{$\pm$0.92} \\
    & & Beam Search & 77.94\scriptsize{$\pm$1.06} & 71.27\scriptsize{$\pm$1.12} & 93.64\scriptsize{$\pm$0.52} & 80.93\scriptsize{$\pm$0.75} \\
    & & DoLa & 71.03\scriptsize{$\pm$1.44} & 63.89\scriptsize{$\pm$1.19} & 96.81\scriptsize{$\pm$0.34} & 76.97\scriptsize{$\pm$0.91} \\
    & & OPERA & 73.96\scriptsize{$\pm$1.39} & 66.79\scriptsize{$\pm$1.25} & 95.36\scriptsize{$\pm$0.54} & 78.55\scriptsize{$\pm$0.95} \\
    & & VCD & 68.98\scriptsize{$\pm$1.03} & 62.59\scriptsize{$\pm$0.85} & 94.41\scriptsize{$\pm$0.42} & 75.27\scriptsize{$\pm$0.62} \\
    & & HALC & 71.03\scriptsize{$\pm$1.44} & 63.89\scriptsize{$\pm$1.19} & 96.81\scriptsize{$\pm$0.34} & 76.97\scriptsize{$\pm$0.91} \\
    & & RVCD & 88.94\scriptsize{$\pm$0.68} & 90.43\scriptsize{$\pm$0.76} & 87.09\scriptsize{$\pm$0.82} & 88.73\scriptsize{$\pm$0.69} \\
    \cmidrule{2-7}
    & \multirow{7}{*}{MiniGPT-4} 
    & Greedy & 56.5\scriptsize{$\pm$1.35} & 53.69\scriptsize{$\pm$0.84} & 94.68\scriptsize{$\pm$0.37} & 68.52\scriptsize{$\pm$0.65} \\
    & & Beam Search & 63.24\scriptsize{$\pm$1.31} & 58.59\scriptsize{$\pm$0.98} & 90.40\scriptsize{$\pm$0.31} & 71.09\scriptsize{$\pm$0.77} \\
    & & DoLa & 70.23\scriptsize{$\pm$0.61} & 71.62\scriptsize{$\pm$1.42} & 67.10\scriptsize{$\pm$1.58} & 69.27\scriptsize{$\pm$0.51} \\
    & & OPERA & 64.03\scriptsize{$\pm$0.99} & 59.12\scriptsize{$\pm$0.77} & 91.02\scriptsize{$\pm$0.41} & 71.68\scriptsize{$\pm$0.55} \\
    & & VCD & 59.80\scriptsize{$\pm$0.99} & 57.13\scriptsize{$\pm$0.81} & 78.61\scriptsize{$\pm$0.90} & 66.17\scriptsize{$\pm$0.68} \\
    & & HALC & 70.23\scriptsize{$\pm$0.61} & 71.62\scriptsize{$\pm$1.42} & 67.10\scriptsize{$\pm$1.58} & 69.27\scriptsize{$\pm$0.51} \\
    & & RVCD & 86.87\scriptsize{$\pm$0.55} & 89.65\scriptsize{$\pm$1.03} & 83.37\scriptsize{$\pm$0.44} & 86.39\scriptsize{$\pm$0.51} \\
    \cmidrule{2-7}
    & \multirow{7}{*}{mPLUG-Owl2} 
    & Greedy & 73.2\scriptsize{$\pm$1.15} & 65.77\scriptsize{$\pm$1.04} & 96.8\scriptsize{$\pm$0.37} & 78.32\scriptsize{$\pm$0.73} \\
    & & Beam Search & 80.02\scriptsize{$\pm$0.70} & 73.67\scriptsize{$\pm$0.83} & 93.42\scriptsize{$\pm$0.25} & 82.38\scriptsize{$\pm$0.51} \\
    & & DoLa & 73.42\scriptsize{$\pm$1.13} & 65.99\scriptsize{$\pm$1.04} & 96.72\scriptsize{$\pm$0.32} & 78.45\scriptsize{$\pm$0.72} \\
    & & OPERA & 78.40\scriptsize{$\pm$0.64} & 71.54\scriptsize{$\pm$0.73} & 94.34\scriptsize{$\pm$0.52} & 81.37\scriptsize{$\pm$0.45} \\
    & & VCD & 72.87\scriptsize{$\pm$0.44} & 66.43\scriptsize{$\pm$0.53} & 92.49\scriptsize{$\pm$0.56} & 77.32\scriptsize{$\pm$0.22} \\
    & & HALC & 73.42\scriptsize{$\pm$1.12} & 65.99\scriptsize{$\pm$1.04} & 96.72\scriptsize{$\pm$0.28} & 78.45\scriptsize{$\pm$0.71} \\
    & & RVCD & 87.82\scriptsize{$\pm$0.91} & 88.50\scriptsize{$\pm$1.64} & 86.97\scriptsize{$\pm$0.26} & 87.72\scriptsize{$\pm$0.80} \\
    \bottomrule
    \end{tabularx}%
    }

\end{table*}


\begin{table*}[t]
    \centering
    \caption{Comparison of the mean of five POPE results on MSCOCO dataset with different decoding baselines under the ‘adversarial’ settings. Higher accuracy (Acc.), precision (Prec.), and F$_1$ score indicate better performance.}
    \label{tab:pope_all_2}
    \resizebox{\textwidth}{!}{
    \begin{tabularx}{\textwidth}{@{}p{2cm}p{2.5cm}p{2cm}XXXX@{}}
    \toprule
    Setting & Model & Decoding & Accuracy & Precision & Recall & $F_{1}$ Score \\
    \midrule

    \multirow{21}{*}{Adversarial} 
    & \multirow{7}{*}{LLaVA-1.5} 
    & Greedy & 65.92\scriptsize{$\pm$0.88} & 59.84\scriptsize{$\pm$0.65} & 96.82\scriptsize{$\pm$0.37} & 73.97\scriptsize{$\pm$0.51} \\
    & & Beam Search & 73.23\scriptsize{$\pm$0.90} & 66.50\scriptsize{$\pm$0.77} & 93.64\scriptsize{$\pm$0.52} & 77.77\scriptsize{$\pm$0.66} \\
    & & DoLa & 66.20\scriptsize{$\pm$0.93} & 60.05\scriptsize{$\pm$0.69} & 96.81\scriptsize{$\pm$0.34} & 74.12\scriptsize{$\pm$0.54} \\
    & & OPERA & 69.98\scriptsize{$\pm$1.00} & 63.26\scriptsize{$\pm$0.78} & 95.36\scriptsize{$\pm$0.54} & 76.06\scriptsize{$\pm$0.67} \\
    & & VCD & 66.31\scriptsize{$\pm$0.27} & 60.42\scriptsize{$\pm$0.18} & 94.56\scriptsize{$\pm$0.37} & 73.73\scriptsize{$\pm$0.21} \\
    & & HALC & 66.20\scriptsize{$\pm$0.93} & 60.05\scriptsize{$\pm$0.69} & 96.81\scriptsize{$\pm$0.34} & 74.12\scriptsize{$\pm$0.54} \\
    & & RVCD & 85.36\scriptsize{$\pm$0.61} & 84.17\scriptsize{$\pm$0.73} & 87.09\scriptsize{$\pm$0.82} & 85.61\scriptsize{$\pm$0.60} \\
    \cmidrule{2-7}
    & \multirow{7}{*}{MiniGPT-4} 
    & Greedy & 56.72\scriptsize{$\pm$0.76} & 53.82\scriptsize{$\pm$0.48} & 94.68\scriptsize{$\pm$0.37} & 68.63\scriptsize{$\pm$0.31} \\
    & & Beam Search & 61.70\scriptsize{$\pm$1.33} & 57.45\scriptsize{$\pm$0.97} & 90.40\scriptsize{$\pm$0.31} & 70.25\scriptsize{$\pm$0.72} \\
    & & DoLa & 69.00\scriptsize{$\pm$0.89} & 69.76\scriptsize{$\pm$1.20} & 67.10\scriptsize{$\pm$1.58} & 68.39\scriptsize{$\pm$0.95} \\
    & & OPERA & 61.95\scriptsize{$\pm$1.69} & 57.57\scriptsize{$\pm$1.24} & 91.02\scriptsize{$\pm$0.41} & 70.53\scriptsize{$\pm$0.91} \\
    & & VCD & 59.32\scriptsize{$\pm$1.15} & 56.73\scriptsize{$\pm$0.91} & 78.61\scriptsize{$\pm$0.90} & 65.90\scriptsize{$\pm$0.76} \\
    & & HALC & 69.00\scriptsize{$\pm$0.89} & 69.76\scriptsize{$\pm$1.20} & 67.10\scriptsize{$\pm$1.58} & 68.39\scriptsize{$\pm$0.95} \\
    & & RVCD & 83.06\scriptsize{$\pm$1.36} & 82.92\scriptsize{$\pm$2.43} & 83.37\scriptsize{$\pm$0.44} & 83.13\scriptsize{$\pm$1.10} \\
    \cmidrule{2-7}
    & \multirow{7}{*}{mPLUG-Owl2} 
    & Greedy & 68.20\scriptsize{$\pm$1.78} & 61.60\scriptsize{$\pm$1.38} & 96.80\scriptsize{$\pm$0.37} & 75.28\scriptsize{$\pm$1.05} \\
    & & Beam Search & 74.28\scriptsize{$\pm$0.77} & 67.56\scriptsize{$\pm$0.79} & 93.42\scriptsize{$\pm$0.25} & 78.41\scriptsize{$\pm$0.49} \\
    & & DoLa & 68.42\scriptsize{$\pm$1.77} & 61.78\scriptsize{$\pm$1.38} & 96.72\scriptsize{$\pm$0.32} & 75.39\scriptsize{$\pm$1.04} \\
    & & OPERA & 72.75\scriptsize{$\pm$1.06} & 65.90\scriptsize{$\pm$0.95} & 94.34\scriptsize{$\pm$0.52} & 77.59\scriptsize{$\pm$0.70} \\
    & & VCD & 69.65\scriptsize{$\pm$0.76} & 63.46\scriptsize{$\pm$0.64} & 92.66\scriptsize{$\pm$0.21} & 75.33\scriptsize{$\pm$0.47} \\
    & & HALC & 68.40\scriptsize{$\pm$1.75} & 61.77\scriptsize{$\pm$1.37} & 96.72\scriptsize{$\pm$0.28} & 75.38\scriptsize{$\pm$1.03} \\
    & & RVCD & 85.48\scriptsize{$\pm$0.41} & 84.46\scriptsize{$\pm$0.64} & 86.97\scriptsize{$\pm$0.26} & 85.70\scriptsize{$\pm$0.36} \\
    
    \bottomrule
    \end{tabularx}%
    }
\end{table*}

\section{MME Experiment Details}
\label{sec:appendix}
For reliability, in MME evaluations, instead of using the offline approach employed in previous studies~\cite{chen2024halc, zhuang2024game}, we utilized a query concatenating the prompt "Please describe this image and then answer the question. " with the original questions from MME to enable automated evaluation. Subsequently, we assessed whether positive/negative words were present in the output captions. Detailed information is provided in Table~\ref{tab:mme_details}.

\begin{center}
\captionof{table}{Comparison of Decoding Methods Performances on MME Sub-tasks: Existence, Color, Count, Position.}
\label{tab:decoder_comparison}
\resizebox{0.5\textwidth}{!}{%
\begin{tabularx}{\textwidth}{@{}lXXXXXXXX@{}}
\toprule
Model & Decoder & Existence & Color & Count & Position & Tokens & Samples \\
\midrule
\multirow{6}{*}{LLaVA-1.5} 
& RVCD       & 123.33 & 95.0 & 60.0 & 65.0 & 128 & 120 \\
& Greedy     & 105.0 & 65.0 & 60.0 & 50.0 & 128 & 120 \\
& VCD        & 100.0 & 85.0 & 56.66 & 60.0 & 128 & 120 \\
& Opera      & 88.33 & 80.0 & 60.0 & 53.33 & 128 & 120 \\
& DoLa       & 75.0 & 60.0 & 53.33 & 55.00 & 128 & 120 \\
& HALC       & 73.33 & 65.0 & 58.33 & 53.33 & 128 & 120 \\
\midrule
\multirow{6}{*}{MiniGPT-4} 
& RVCD       & 130.0 & 70.0 & 51.66 & 53.33 & 128 & 120 \\
& Greedy     & 108.33 & 78.33 & 48.33 & 56.66 & 128 & 120 \\
& VCD        & 125.0 & 65.0 & 53.33 & 51.66 & 128 & 120 \\
& Opera      & 71.66 & 53.33 & 46.66 & 56.66 & 128 & 120 \\
& DoLa       & 128.33 & 78.33 & 60.0 & 60.0 & 128 & 120 \\
& HALC       & 138.33 & 78.33 & 63.33 & 58.33 & 128 & 120 \\
\midrule
\multirow{6}{*}{mPLUG-Owl2} 
& RVCD       & 130.0 & 125.0 & 70.0 & 53.33 & 128 & 120 \\
& Greedy     & 111.66 & 120.0 & 68.33 & 70.0 & 128 & 120 \\
& VCD        & 126.66 & 91.66 & 81.66 & 55.0 & 128 & 120 \\
& Opera      & 105.0 & 90.0 & 63.33 & 50.0 & 128 & 120 \\
& DoLa       & 111.66 & 120.0 & 65.0 & 65.0 & 128 & 120 \\
& HALC       & 98.33 & 115.0 & 65.0 & 53.33 & 128 & 120 \\
\bottomrule
\end{tabularx}%
}
\label{tab:mme_details}
\end{center}


\section{Token Probabilities of Single Concept Images}
\label{sec:appendixF}
The bar graph depicted in Figure~\ref{fig:top5_co_occur} visualizes the probability distribution of the first token when LVLM describes single concept images with the prompt, "What is this? Answer in one word." The probabilities are min-max scaled for better visualization. Tokens that start with an underbar represent the first token, and the subsequent remaining tokens forming the word are represented in italic with an underline. The remaining tokens forming the words in Figure~\ref{fig:top5_co_occur} were inferred through next-token prediction, where the LVLM's input consists of the prompt followed by the first token of the word. Table~\ref{tab:entity_probabilities} represents the original output token probabilities for the first tokens of example images depicted in Figure~\ref{fig:top5_co_occur}.


\begin{table*}[ht]
\centering
\resizebox{1\linewidth}{!}{ 
\begin{tabular}{lccccc}
\toprule
\textbf{Image} & \textbf{Token 1} & \textbf{Token 2} & \textbf{Token 3} & \textbf{Token 4} & \textbf{Token 5} \\
\midrule
Toilet    & 0.9600 (\textbf{\textunderscore To}\uline{\textit{ilet}}) & 0.0073 (\textbf{\textunderscore Bath}) & 0.0067 (\textbf{\textunderscore L}\uline{\textit{id}}) & 0.0034 (\textbf{\textunderscore Bowl}) & 0.0020 (\textbf{\textunderscore Pot}\uline{\textit{ty}}) \\
Cat       & 0.9692 (\textbf{\textunderscore Cat}) & 0.0073 (\textbf{\textunderscore K}\uline{\textit{itten}}) & 0.0024 (\textbf{\textunderscore Dog}) & 0.0017 (\textbf{\textunderscore T}\uline{\textit{iger}}) & 0.0015 (\textbf{\textunderscore St}\uline{\textit{riped}}) \\
Fork      & 0.6636 (\textbf{\textunderscore F}\uline{\textit{ork}}) & 0.0971 (\textbf{\textunderscore Kn}\uline{\textit{ife}}) & 0.0745 (\textbf{\textunderscore Ut}\uline{\textit{ensil}}) & 0.0272 (\textbf{\textunderscore Spo}\uline{\textit{on}}) & 0.0231 (\textbf{\textunderscore Table}) \\
Spoon    & 0.9463 (\textbf{\textunderscore Spo}\uline{\textit{on}}) & 0.0100 (\textbf{\textunderscore S}\uline{\textit{poon}}) & 0.0063 (\textbf{\textunderscore Sp}\uline{\textit{oon}}) & 0.0062 (\textbf{\textunderscore Silver}) & 0.0056 (\textbf{\textunderscore Bowl}) \\
Keyboard & 0.9419 (\textbf{\textunderscore Key}\uline{\textit{board}}) & 0.0349 (\textbf{\textunderscore Computer}) & 0.0031 (\textbf{\textunderscore Ke}\uline{\textit{yboard}}) & 0.0018 (\textbf{\textunderscore Mouse}) & 0.0008 (\textbf{\textunderscore keyboard}) \\
\bottomrule
\end{tabular}
}
\caption{Original output probabilities mapped to first tokens before min-max scaling. Token 1$\sim$5 represents top-5 first tokens of the given single concept image. Bolded parts represent the first tokens, and the subsequent remaining tokens forming the word are represented in italic with an underline.}
\label{tab:entity_probabilities}
\end{table*}

\section{Hyperparameters settings}
\label{sec:appendixG}

The hyperparameter settings of RVCD are shown in Table~\ref{tab:hyperparams}. The experimental configurations for other decoding methods with evaluation of CHAIR and BLEU scores based on natural language processing were aligned with the detailed hyperparameter settings and evaluation settings specified in HALC~\cite{chen2024halc}.

Unlike previous studies, we clarify that RVCD does not adopt the adaptive plausibility threshold. Contrastive decoding-based prior studies propose an adaptive plausibility threshold~\cite{chen2024halc, leng2023visual, chuang2023dola} to mitigate situations where their method promotes the generation of implausible tokens. However, we find that this approach does not provide significant benefits when applied to RVCD. Since our RVCD already achieves high-quality outputs with state-of-the-art performance without relying on this additional condition, we determine that incorporating it is unnecessary.

We selected YOLO as our object detection model for several key reasons. First, it is both lightweight and computationally efficient~\cite{redmon2015yolo}. Second, because RVCD only requires identifying which objects are present in the input image, the prompt understanding of Grounding DINO~\cite{liu2023groundingdino} or the precise segmentation capabilities of Grounded-SAM~\cite{ren2024groundedsam} are not essential, each utilized in respective previous decoding-based studies~\cite{chen2024halc, zhuang2024game}. Third, to investigate the influence of detection model improvements on RVCD performance, we needed multiple versions of the same model. YOLO has been a cornerstone in the deep learning community for nearly a decade, offering a wide range of open-source releases that made it particularly well-suited to our study. For state-of-the-art experiments, we employed YOLOv8x~\cite{yolov8}, while for the detector ablation study, we additionally used YOLOv3~\cite{redmon2018yolov3}. In both cases, the confidence threshold was set to the default value of 0.25.
\begin{table}[H]
\centering
\caption{RVCD Default Hyperparameter Settings}
\resizebox{0.5\textwidth}{!}{%
\begin{tabular}{l|l}
\hline
\textbf{Parameters}                     & \textbf{Value}               \\ \hline
Negative Logits Regulation Factor $\alpha$           & 1                         \\ \hline
Positive Logits Recovery Factor $\beta$                     & 0.1                            \\ \hline
Object Detection Model, Confidence Threshold                               & YOLOv8x~\cite{yolov8}, 0.25                            \\ \hline
\end{tabular}%
}
\label{tab:hyperparams}
\end{table}

\subsection{$\alpha$ and $\beta$ ablation study detail}
\label{sec:alpha_beta}

For statistical details, refer to Table~\ref{tab:alpha_ablation} and Table~\ref{tab:beta_ablation}. As the $\beta$ value increases from 0, the CHAIR$_I$ and CHAIR$_S$ scores improve, whereas the BLEU score peaks at 0.1 and then starts to deteriorate. This indicates that the influence of positive logits should not be excessive. Therefore, we set $\alpha$ and $\beta$ to 1 and 0.1, respectively. We propose this as the default setting for RVCD.

\begin{table*}[t]
\centering
\small
\captionsetup{aboveskip=5pt, belowskip=-10pt}
\setlength{\tabcolsep}{4.2pt} 
\renewcommand{\arraystretch}{1.2} 
\resizebox{\linewidth}{!}{ 
\begin{tabularx}{\textwidth}{l|ccc|ccc|ccc}
\toprule
\multirow{2}{*}{\raisebox{-0.2em}{\textbf{$\alpha$}}} & \multicolumn{3}{c|}{\textbf{LLaVA-1.5}} & \multicolumn{3}{c|}{\textbf{MiniGPT-4}} & \multicolumn{3}{c}{\textbf{mPLUG-Owl2}} \\
\cline{2-4} \cline{5-7} \cline{8-10}
 
 & CHAIR$_S$ ↓ & CHAIR$_I$ ↓ & BLEU ↑ &  CHAIR$_S$ ↓ & CHAIR$_I$ ↓ & BLEU ↑ & CHAIR$_S$ ↓ & CHAIR$_I$ ↓ & BLEU ↑ \\
\midrule

0.25 & 12.20\scriptsize{$\pm$2.00} & 3.91\scriptsize{$\pm$0.50} &  16.04\scriptsize{$\pm$0.18} & 13.28\scriptsize{$\pm$2.21} & 4.82\scriptsize{$\pm$1.00} & 16.17\scriptsize{$\pm$0.27} & 13.48\scriptsize{$\pm$1.62} & 4.85\scriptsize{$\pm$0.38} & 15.33\scriptsize{$\pm$0.20} \\ 
0.5 & 11.52\scriptsize{$\pm$0.86} & 3.66\scriptsize{$\pm$0.19} &  15.78\scriptsize{$\pm$0.21} & 11.60\scriptsize{$\pm$1.21} & 4.21\scriptsize{$\pm$0.38} & 16.06\scriptsize{$\pm$0.21} & 11.68\scriptsize{$\pm$1.02} & 3.89\scriptsize{$\pm$0.27} & 15.13\scriptsize{$\pm$0.26} \\ 
0.75 & 11.12\scriptsize{$\pm$0.39} & 3.66\scriptsize{$\pm$0.19} &  15.48\scriptsize{$\pm$0.13} & 9.76\scriptsize{$\pm$1.05} & 3.60\scriptsize{$\pm$0.32} & 16.19\scriptsize{$\pm$0.16} & 11.44\scriptsize{$\pm$0.50} & 4.07\scriptsize{$\pm$0.25} & 14.90\scriptsize{$\pm$0.19} \\ 
1.0 & 11.08\scriptsize{$\pm$1.15} & 3.74\scriptsize{$\pm$0.33} &  15.31\scriptsize{$\pm$0.13} & 9.28\scriptsize{$\pm$0.64} & 3.68\scriptsize{$\pm$0.32} & 15.98\scriptsize{$\pm$0.33} & 10.92\scriptsize{$\pm$1.83} & 3.84\scriptsize{$\pm$0.39} & 14.64\scriptsize{$\pm$0.22} \\ 
\midrule
\end{tabularx}
} 
\caption{Ablation study on the $\alpha$ settings for CHAIR$_S$, CHAIR$_I$, and BLEU metrics. The mean and standard deviation of
five samples, each consisting of 500 instances, sampled
with replacement from the MSCOCO 2014 validation
dataset.}
\label{tab:alpha_ablation}
\end{table*}

\begin{table*}[t]
\centering
\small
\captionsetup{aboveskip=5pt, belowskip=-10pt}
\setlength{\tabcolsep}{3.75pt} 
\renewcommand{\arraystretch}{1.2} 
\resizebox{\linewidth}{!}{ 
\begin{tabularx}{\textwidth}{l|ccc|ccc|ccc}
\toprule
\multirow{2}{*}{\raisebox{-0.2em}{\textbf{$\beta$}}} & \multicolumn{3}{c|}{\textbf{LLaVA-1.5}} & \multicolumn{3}{c|}{\textbf{MiniGPT-4}} & \multicolumn{3}{c}{\textbf{mPLUG-Owl2}} \\
\cline{2-4} \cline{5-7} \cline{8-10}
 
 & CHAIR$_S$ ↓ & CHAIR$_I$ ↓ & BLEU ↑ &  CHAIR$_S$ ↓ & CHAIR$_I$ ↓ & BLEU ↑ & CHAIR$_S$ ↓ & CHAIR$_I$ ↓ & BLEU ↑ \\
\midrule

0 & 11.08\scriptsize{$\pm$1.15} & 3.74\scriptsize{$\pm$0.33} &  15.31\scriptsize{$\pm$0.13} & 9.28\scriptsize{$\pm$0.64} & 3.68\scriptsize{$\pm$0.32} & 15.98\scriptsize{$\pm$0.33} & 10.92\scriptsize{$\pm$1.83} & 3.84\scriptsize{$\pm$0.39} & 14.64\scriptsize{$\pm$0.22} \\ 
0.0001 & 11.08\scriptsize{$\pm$1.14} & 3.73\scriptsize{$\pm$0.37} &  15.32\scriptsize{$\pm$0.13} & 9.40\scriptsize{$\pm$0.65} & 3.69\scriptsize{$\pm$0.32} & 16.00\scriptsize{$\pm$0.31} & 10.96\scriptsize{$\pm$1.80} & 3.86\scriptsize{$\pm$0.39} & 14.64\scriptsize{$\pm$0.21} \\ 
0.001 & 11.20\scriptsize{$\pm$1.09} & 3.76\scriptsize{$\pm$0.32} &  15.32\scriptsize{$\pm$0.13} & 9.36\scriptsize{$\pm$0.71} & 3.65\scriptsize{$\pm$0.30} & 16.00\scriptsize{$\pm$0.31} & 11.00\scriptsize{$\pm$1.81} & 3.86\scriptsize{$\pm$0.45} & 14.64\scriptsize{$\pm$0.20} \\ 

0.1 & 10.84\scriptsize{$\pm$0.89} & 3.70\scriptsize{$\pm$0.41} &  15.45\scriptsize{$\pm$0.16} & 8.64\scriptsize{$\pm$0.95} & 3.32\scriptsize{$\pm$0.36} & 16.13\scriptsize{$\pm$0.14} & 10.52\scriptsize{$\pm$1.55} & 3.63\scriptsize{$\pm$0.34} & 14.78\scriptsize{$\pm$0.21} \\ 

0.3 & 9.28\scriptsize{$\pm$0.76} & 3.20\scriptsize{$\pm$0.34} &  15.39\scriptsize{$\pm$0.21} & 9.52\scriptsize{$\pm$1.19} & 3.56\scriptsize{$\pm$0.32} & 15.87\scriptsize{$\pm$0.20} & 10.48\scriptsize{$\pm$1.51} & 3.64\scriptsize{$\pm$0.35} & 14.74\scriptsize{$\pm$0.21} \\ 

0.5 & 8.24\scriptsize{$\pm$0.83} & 3.01\scriptsize{$\pm$0.40} &  14.90\scriptsize{$\pm$0.20} & 7.96\scriptsize{$\pm$0.59} & 3.15\scriptsize{$\pm$0.32} & 15.33\scriptsize{$\pm$0.05} & 10.20\scriptsize{$\pm$0.66} & 3.48\scriptsize{$\pm$0.41} & 14.37\scriptsize{$\pm$0.18} \\ 

0.7 & 7.04\scriptsize{$\pm$1.16} & 2.65\scriptsize{$\pm$0.53} &  14.33\scriptsize{$\pm$0.24} & 7.76\scriptsize{$\pm$0.67} & 3.49\scriptsize{$\pm$0.44} & 14.69\scriptsize{$\pm$0.19} & 9.12\scriptsize{$\pm$0.91} & 3.60\scriptsize{$\pm$0.45} & 13.85\scriptsize{$\pm$0.17} \\ 

0.9 & 7.44\scriptsize{$\pm$0.89} & 3.11\scriptsize{$\pm$0.37} &  13.92\scriptsize{$\pm$0.21} & 7.24\scriptsize{$\pm$0.85} & 3.60\scriptsize{$\pm$0.59} & 14.12\scriptsize{$\pm$0.24} & 8.76\scriptsize{$\pm$1.51} & 3.59\scriptsize{$\pm$0.46} & 13.39\scriptsize{$\pm$0.26} \\ 

\midrule
\end{tabularx}
} 
\caption{Ablation study on the $\beta$ settings for CHAIR$_S$, CHAIR$_I$, and BLEU metrics. The mean and standard deviation of
five samples, each consisting of 500 instances, sampled
with replacement from the MSCOCO 2014 validation
dataset.}
\label{tab:beta_ablation}
\end{table*}

\vspace{\baselineskip} 
 
\section{Experiment Results on LLaVA-Bench.}
\label{sec:appendix_llavabench}
\begin{figure*}[htbp]
    \centering
    \includegraphics[width=\textwidth,height=0.9\textheight,keepaspectratio]{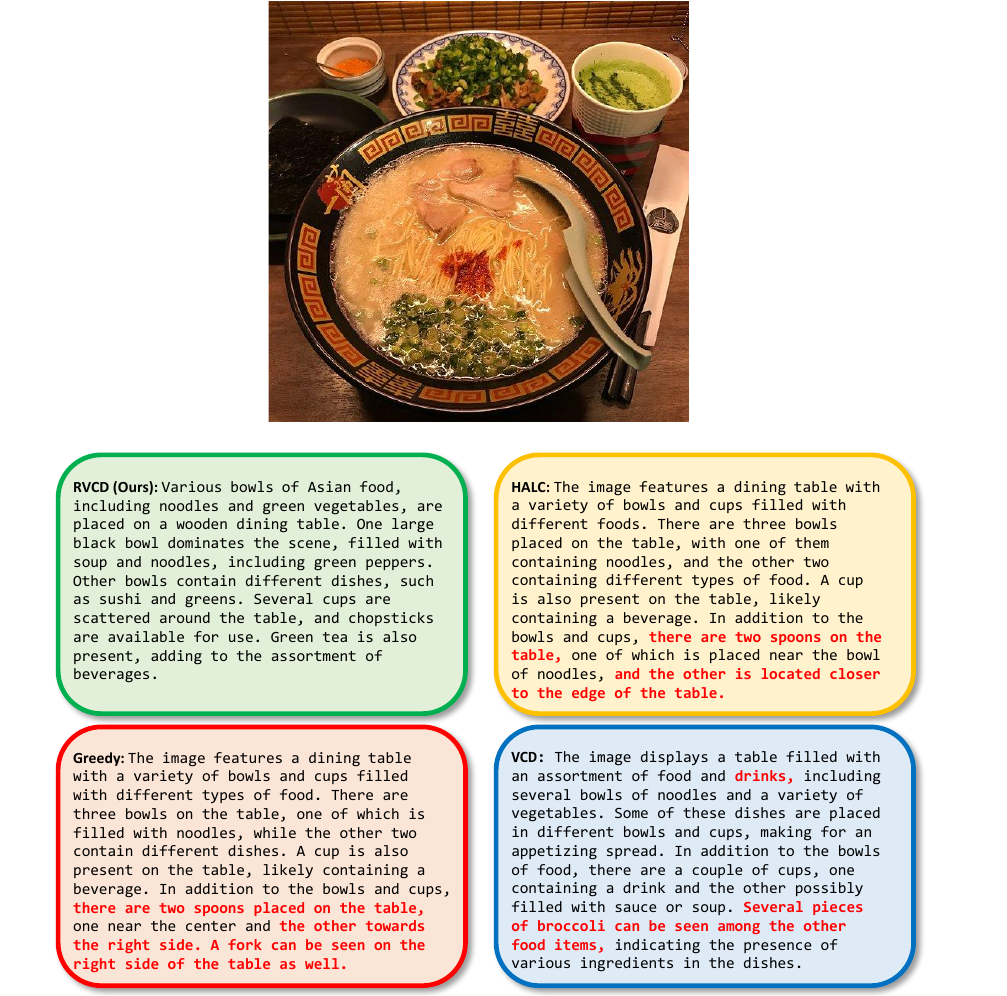}
    \caption{LLaVA-Bench results comparing our RVCD and other methods with LLaVA-1.5 backbone.}
    \label{fig:example_llava}
\end{figure*}

\begin{figure*}[htbp]
    \centering
    \includegraphics[width=\textwidth,height=0.9\textheight,keepaspectratio]{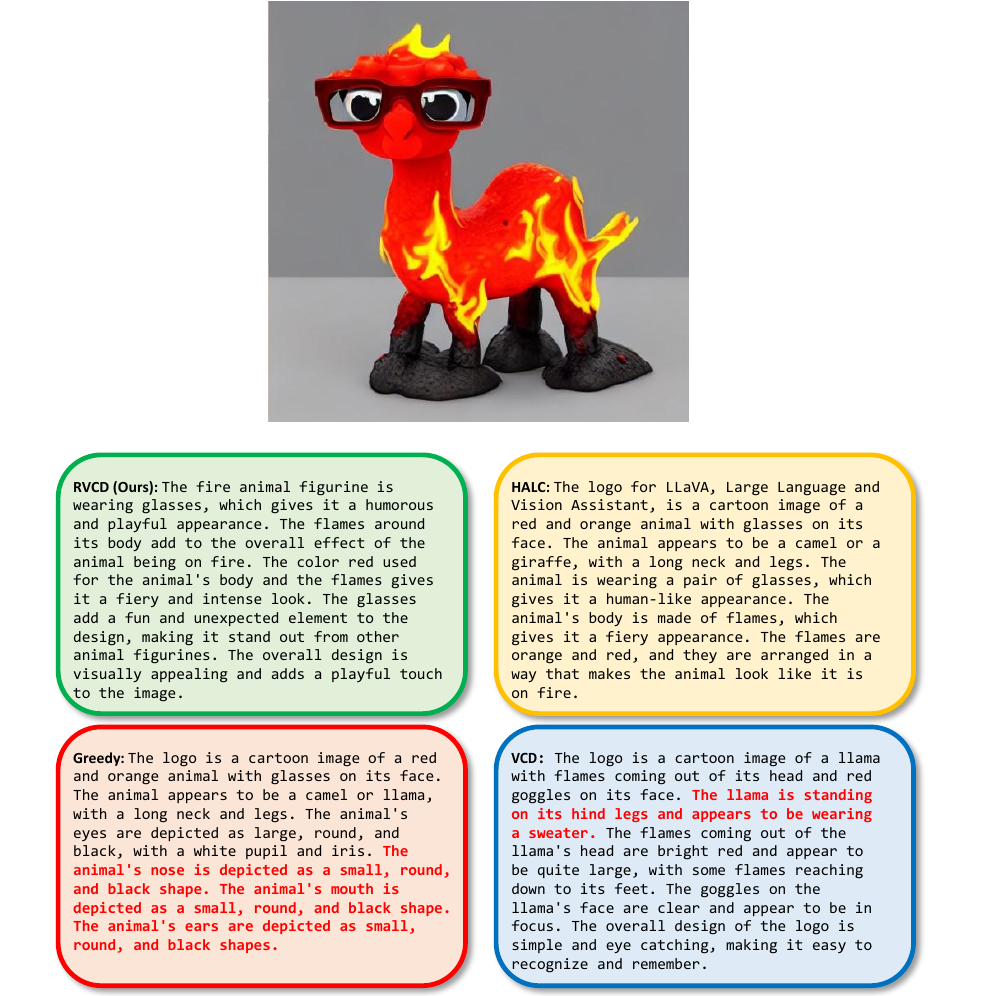}
    \caption{LLaVA-Bench results comparing our RVCD and other methods with MiniGPT-4 backbone.}
    \label{fig:example_minigpt4}
\end{figure*}

\begin{figure*}[htbp]
    \centering
    \includegraphics[width=\textwidth,height=0.9\textheight,keepaspectratio]{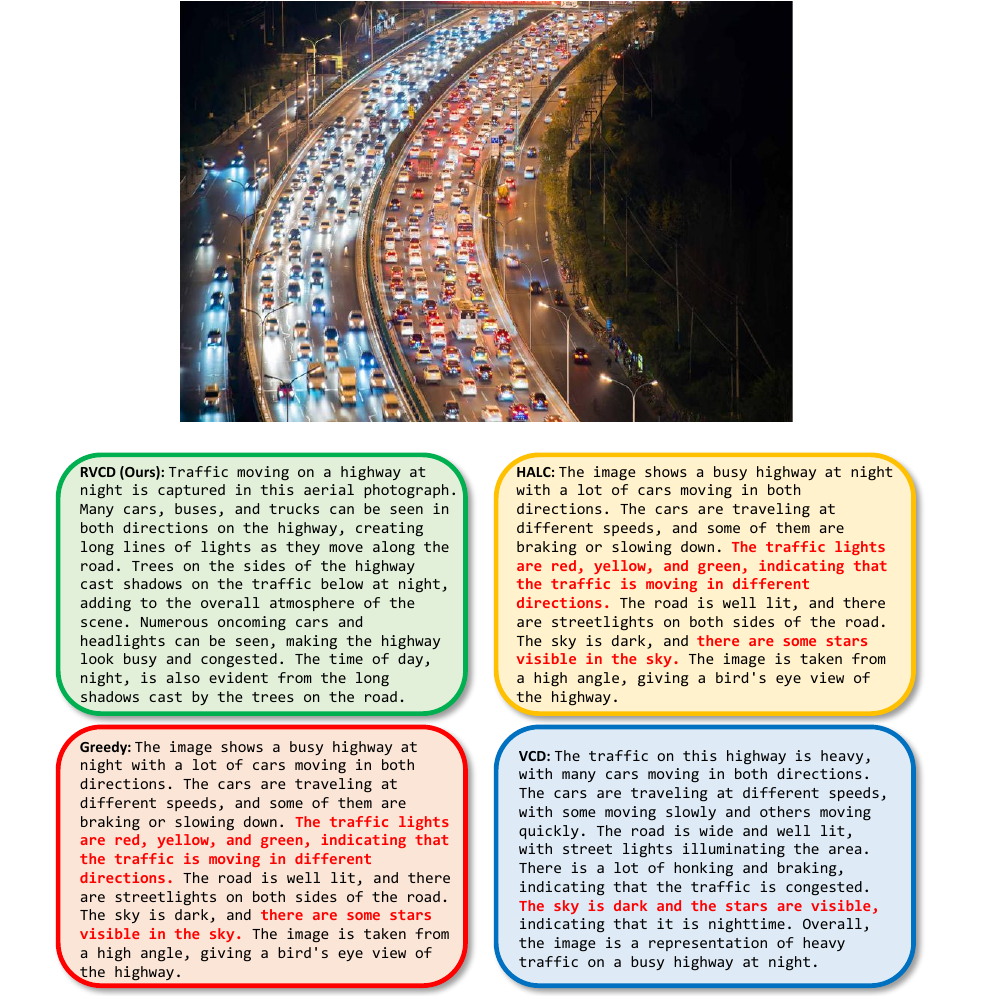}
    \caption{LLaVA-Bench results comparing our RVCD and other methods with MiniGPT-4 backbone.}
    \label{fig:example_minigpt4_2}
\end{figure*}

\begin{figure*}[htbp]
    \centering
    \includegraphics[width=\textwidth,height=0.9\textheight,keepaspectratio]{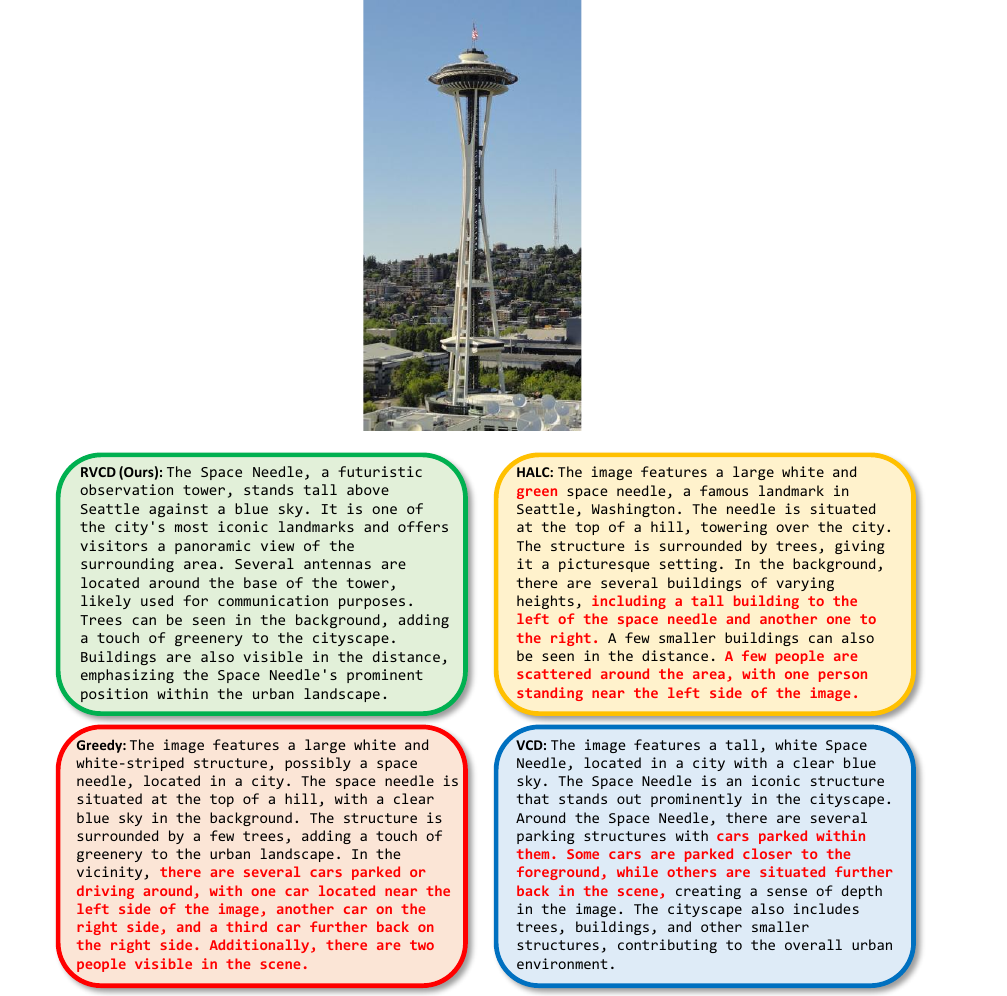}
    \caption{LLaVA-Bench results comparing our RVCD and other methods with mPLUG-Owl2 backbone.}
    \label{fig:example_owl2}
\end{figure*}

We utilize LLaVA-Bench~\cite{liu2023visual2} for qualitative case studies (Figure~\ref{fig:example_llava}, Figure~\ref{fig:example_minigpt4},
Figure~\ref{fig:example_minigpt4_2},
Figure~\ref{fig:example_owl2}). Captions generated by RVCD and other decoding methods designed to mitigate OH are presented, with red fonts indicating occurrences of OH and highlighting cases where hallucinations occurred in object existence, attributes, or relationships. The identical examples of HALC and Greedy in Figure~\ref{fig:example_minigpt4_2} are actual decoded result.

\clearpage

\end{document}